\documentclass[lettersize,journal]{IEEEtran}
\usepackage{amsmath,amsfonts}
\usepackage{algorithmic}
\usepackage{algorithm}
\usepackage{array}
\usepackage[caption=false,font=normalsize,labelfont=sf,textfont=sf]{subfig}
\usepackage{textcomp}
\usepackage{stfloats}
\usepackage{url}
\usepackage{verbatim}
\usepackage{graphicx}
\usepackage{cite}
\usepackage{amssymb}
\usepackage{booktabs}
\usepackage{multirow}
\usepackage[colorlinks=true, linkcolor=blue, citecolor=blue, urlcolor=black]{hyperref}
\begin{document}


\title{Structure-Aware Matching for Robust UAV Thermal-to-Satellite Geo-Localization}

\author{Chun Liu, Yuxin Wang, Zhenxiang Huo, Linlin Zhu, Tao Xu, Henggang Han

\thanks{\emph{Corresponding author: Tao Xu.}}
\thanks{Chun Liu, Tao Xu and Zhenxiang Huo are with the State Key Laboratory of Spatial Datum, Henan University, Zhengzhou 450046, China (e-mail: liuchun@henu.edu.cn; txu@henu.edu.cn; hzxcurry@henu.edu.cn;).}
\thanks{Chun Liu and Zhenxiang Huo is also with the College of Remote Sensing and Geoinformatics Engineering, Henan University, Zhengzhou 450046, China (e-mail: liuchun@henu.edu.cn; hzxcurry@henu.edu.cn;).}
\thanks{Yuxin Wang, Linlin Zhu, and Tao Xu are with the School of Computer and Information Engineering, Henan University, Zhengzhou 450046, China (e-mail: yuxinwang@henu.edu.cn; llzhu@henu.edu.cn; txu@henu.edu.cn).}
\thanks{Henggang Han is with Henan Yunhuan NetLink UAV Technology Co., Ltd., Anyang 455000, China (e-mail: hanhenggang@163.com).}

}

\markboth{Journal of \LaTeX\ Class Files,~Vol.~14, No.~8, August~2021}%
{Shell \MakeLowercase{\textit{et al.}}: A Sample Article Using IEEEtran.cls for IEEE Journals}


\maketitle

\begin{abstract}

 UAV thermal-to-satellite  geo-localization localizes UAVs by matching real-time thermal images against geo-referenced satellite images, enabling autonomous navigation in GNSS-denied and low-visibility conditions. However, substantial cross-modal disparities in color, brightness and viewpoint hinder conventional appearance-based methods from establishing reliable correspondences. While existing approaches predominantly address global modality shifts, they overlook the invariant scene structures inherently shared between thermal and satellite views. To bridge this gap, we propose SAMF, a Structure-Aware Matching Framework that explicitly exploits stable cross-modal structural cues to improve   UAV thermal geo-localization. Specifically, the Directional Structure Enhancement Module (DSEM) strengthens robust structural representations by capturing directional and low-frequency spatial patterns, and the Modality-Aware Affine Remapping Module (MARM) subsequently applies affine correction to these enhanced features, further promoting cross-modal alignment. Additionally, we introduce the Thermal-UAV-Paired benchmark derived from the Thermal-UAV dataset. Extensive experiments on both the Boson-nighttime and Thermal-UAV-Paired datasets show that SAMF achieves state-of-the-art performance. Ablation studies further validate the effectiveness and synergistic complementarity of DSEM and MARM. The code and dataset are released at: \url{https://github.com/YuxinWang879/SAMF.git}.

\end{abstract}


\section{Introduction}
\IEEEPARstart{R}{eliable} geo-localization is essential for autonomous UAV flight and mission execution. However, in extreme environments involving complex electromagnetic confrontation or geographic occlusion, Global Navigation Satellite System (GNSS) signals may be blocked, interfered with or spoofed, making it difficult for UAVs to continuously obtain reliable position information \cite{ref1,ref2}. To reduce dependence on GNSS, cross-view geo-localization determines the UAV’s location by matching images captured in real time by the UAV with geo-tagged satellite images \cite{ref3}.

In recent years, UAV-to-satellite cross-view geo-localization has received increasing  attention. Existing studies mainly focus on daytime visible-light scenarios and improve the matching capability between UAV images and satellite images through feature representation, local information modeling or cross-view alignment \cite{ref6,ref7,ref8,ref9}. However, when UAVs operate at night or under low-light conditions, visible-light images often fail to preserve sufficient scene details. In contrast, thermal images are less affected by illumination variations and can continuously capture ground scenes.  Therefore, cross-view matching between UAV thermal  images and satellite images has emerged as a  pivotal problem for UAV autonomous localization in low-visibility conditions\cite{ref4}.

Due to significant differences in color, brightness, and texture between thermal and satellite images  \cite{ref4}, even if they correspond to the same geographic location, their appearance may differ greatly. The above differences, combined with variations in viewpoint, scale, and direction, make it highly challenging to establish reliable matching relationships between two types of images. Current methods for UAV thermal-to-satellite geo-localization can be broadly categorized into two groups according to how they address the modality discrepancy between thermal and satellite images. The first category focuses on reducing the modality distribution gap between thermal and satellite images, mainly through domain-adversarial training \cite{ref10}, conditional-GAN-based image synthesis \cite{ref11}, and sparse generative models \cite{ref12}. For example, Xiao et al. \cite{ref4} integrated these techniques into a thermal geo-localization framework and achieved competitive  geo-localization performance on the Boson-nighttime dataset. The second category bypasses feature distribution alignment and instead directly models the geometric transformation between thermal images and satellite images through deep homography estimation. STHN \cite{ref13} estimates a fine-grained homography matrix based on coarse retrieval results to achieve pixel-level geometric alignment, while UASTHN \cite{ref14} further introduces an uncertainty-aware mechanism to automatically identify and discard unreliable matches in geometric estimation.

\begin{figure}[!t] 
\centering
\includegraphics[width=\linewidth]{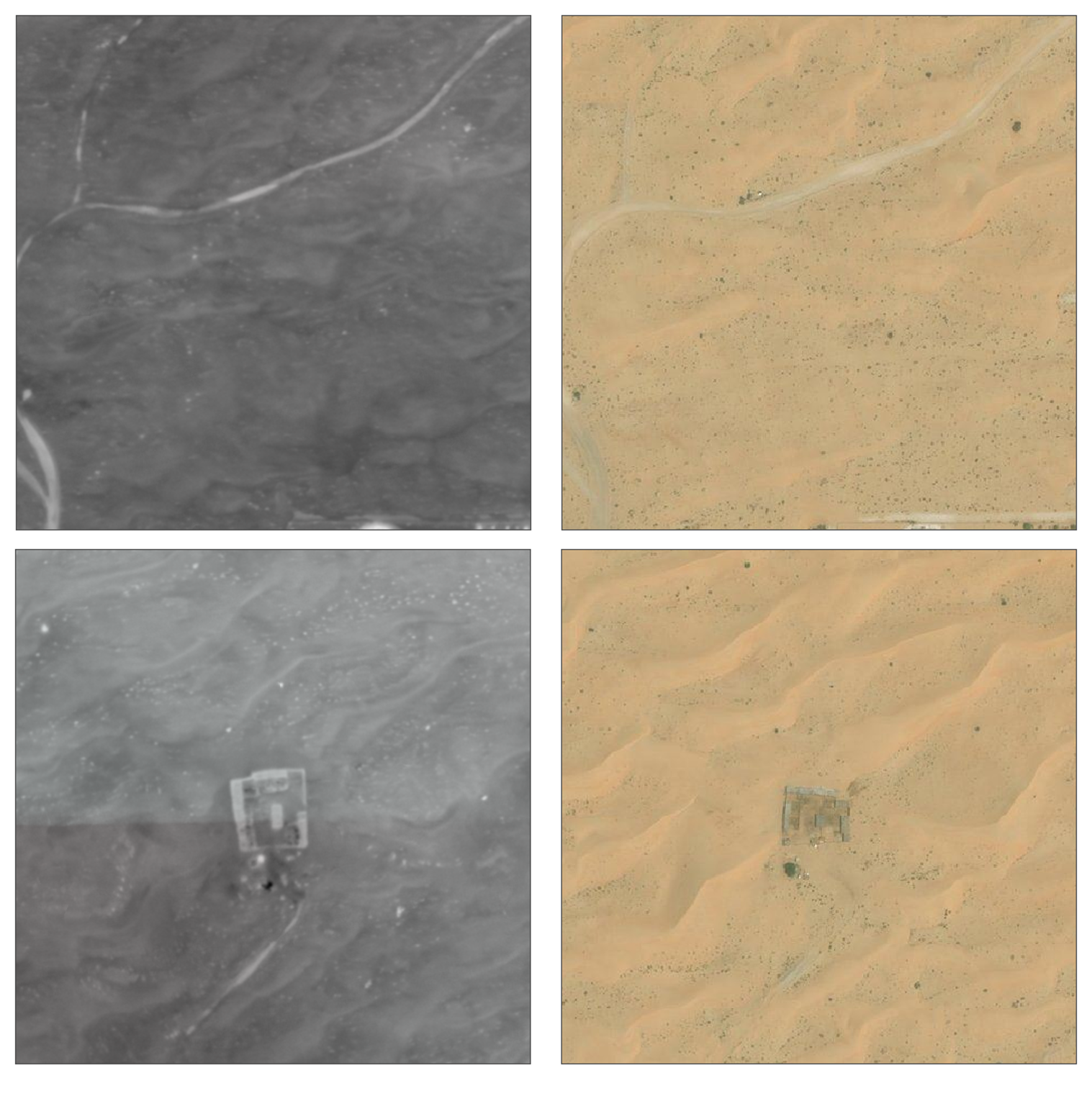}
\caption{Representative examples from the Boson-nighttime dataset. The left column displays UAV thermal images, while the right column presents the corresponding satellite reference images. Although thermal and satellite images captured at the same locations exhibit substantial visual appearance discrepancies, geometric structures such as surface stripe orientations, regional boundary contours, and coarse-grained spatial layouts exhibit strong cross-modal consistency. These consistent structural features provide potentially stable cues for cross-modal matching.}
\label{fig:boson_examples} 
\end{figure}

While existing approaches have advanced the task through distribution alignment and geometric estimation, the intrinsic structural cues shared between satellite and thermal imagery have not been fully exploited. In reality, these cues are pivotal for robust cross-modal matching. As illustrated in Fig.~\ref{fig:boson_examples}, despite significant discrepancies in radiometry, local texture, and brightness, thermal and satellite images preserve consistent large-scale scene structures. Specifically, attributes such as surface orientation, region boundaries, terrain trends, and global spatial layouts exhibit strong structural coherence across modalities. Unlike appearance features that are highly sensitive to sensor characteristics, these structural cues reflect the spatial organization of ground scenes and thus demonstrate superior cross-modal stability \cite{ref15,ref16}.  Therefore, rather than attempting to directly bridge the global modality gap,  leveraging these stable structural cues provides a more effective mechanism to mitigate appearance inconsistencies.

To this end, we propose a Structure-Aware Matching Framework (SAMF) that explicitly models and leverages these stable structural cues to enhance the accuracy and reliability of UAV thermal geo-localization. Specifically, SAMF incorporates a Directional Structure Enhancement Module (DSEM) to reinforce robust structural representations by exploiting directional and low-frequency spatial information. Complementing this, a Modality-Aware Affine Remapping Module (MARM) is introduced to perform modality-specific affine corrections on the structure-enhanced features, further mitigating residual cross-modal discrepancies. We validate SAMF on the Boson-nighttime dataset and the newly constructed Thermal-UAV-Paired dataset. The Boson-nighttime dataset was originally established by Xiao et al. \cite{ref4}. To rigorously evaluate our proposed method, we introduce the Thermal-UAV-Paired dataset, which is derived from the Thermal-UAV dataset \cite{ref17} by establishing precise one-to-one correspondences between thermal UAV images and satellite imagery. Experimental results demonstrate that SAMF achieves state-of-the-art performance on both benchmarks. Furthermore, ablation studies verify the effectiveness and complementarity of the DSEM and MARM modules, confirming that exploiting stable cross-modal structures significantly bolsters geo-localization performance.

The main contributions of this paper are summarized as follows:

\begin{enumerate}
	
	\item We propose a Structure-Aware Matching Framework tailored for  UAV thermal-to-satellite geo-localization. SAMF explicitly leverages stable cross-modal structural cues to substantially improve localization accuracy and reliability under low-visibility conditions.
	
	\item We introduce a Directional Structure Enhancement Module to extract and reinforce invariant spatial structures across modalities. By capturing surface texture orientations, weak boundary trends, and coarse-grained layouts, DSEM ensures robust performance even in weak-texture and low-semantic environments.

	\item 	We design a Modality-Aware Affine Remapping Module to perform lightweight modality alignment on the structure-enhanced features. MARM effectively mitigates cross-modal feature discrepancies while preserving spatial structures, forming a  complementary processing pipeline with DSEM.

	\item 	We construct and release the Thermal-UAV-Paired benchmark based on the Thermal-UAV dataset. Comprehensive evaluations on both the Boson-nighttime and Thermal-UAV-Paired datasets demonstrate the state-of-the-art performance of our proposed framework.
	
\end{enumerate}

The remainder of this paper is organized as follows. Section~\ref{sec:related work} reviews related studies on cross-view UAV geo-localization, cross-modal feature alignment, and spatial structural feature learning. Section~\ref{sec:methodology} introduces the overall framework of SAMF and its main modules. Section~\ref{sec:experiments} presents the experimental settings, comparison results, and ablation analysis. Finally, Section~\ref{sec:conclusions} concludes this paper.

\section{RELATED WORK}
\label{sec:related work}

This section provides an overview of related works covering  cross-view UAV geo-localization, cross-modal feature alignment, and spatial structure feature learning.

\subsection{Cross-View UAV Geo-Localization}

Cross-view geo-localization usually estimates the location of a query image by matching it against a large-scale geo-tagged reference database. Early studies established the task framework for ground-to-satellite and UAV-to-satellite matching, and improved the correspondence between different views through feature representation, feature aggregation, and metric learning \cite{ref3,ref5,ref6,ref7}. Subsequent studies have improved UAV--satellite matching through different mechanisms, including pose estimation with satellite imagery \cite{ref18}, UAV--satellite view synthesis \cite{ref19}, local-pattern modeling \cite{ref20}, feature segmentation and regional alignment \cite{ref21}, and cross-view consistent attention \cite{ref22}. Research on reference-image representation, geography-aware sampling, and benchmark construction has also improved the robustness of localization systems to viewpoint and scene variations \cite{ref23,ref24,ref25}. These methods have advanced UAV--satellite cross-view geo-localization, but most of them are developed for daytime visible-light images. Their matching processes still rely substantially on color, texture, and local appearance cues, whose stability decreases under insufficient illumination and nighttime imaging conditions.

Thermal infrared sensing can capture ground scenes with limited dependence on ambient illumination and therefore provides an alternative observation modality for UAV geo-localization in low-visibility conditions \cite{ref4}. Existing thermal-to-satellite geo-localization studies mainly reduce the difficulty of cross-modal matching from two perspectives. The first line, represented by Xiao et al. \cite{ref4}, reduces the overall distribution discrepancy between thermal and satellite images by combining domain-adversarial training \cite{ref10}, conditional GAN-based image-to-image translation \cite{ref11}, and sparse GAN-based thermal image generation \cite{ref12}, together with data augmentation. The other group estimates the geometric transformation between the two views after coarse retrieval and performs fine-grained localization through homography modeling and local matching \cite{ref13,ref14}. Other studies have used multimodal data augmentation to improve robustness to complex imaging conditions, or combined zero-shot retrieval with local feature matching to extend the applicability of thermal geo-localization \cite{ref15,ref26}. These studies have advanced the task from the perspectives of distribution alignment, geometric estimation, and generalization capability. However, they generally treat cross-modal differences as an overall distribution shift or a geometric transformation, failing to exploit the invariant structural cues shared across thermal and satellite modalities.  This limitation  restricts model performance in weak-texture and low-semantic scenes, where appearance information is scarce and stable structural cues are indispensable.

\subsection{Cross-Modal Feature Alignment}

Reducing feature distribution discrepancies between heterogeneous modalities is a common strategy in cross-modal matching. Domain-adversarial training uses a domain classifier and a gradient reversal mechanism to encourage the feature extractor to learn representations that are difficult to distinguish by modality \cite{ref10}. CyCADA further combines domain adaptation with cycle-consistent image generation to reduce appearance discrepancies between different domains \cite{ref27,ref28}. In thermal imaging, conditional generative models and sparse generative models have been used to establish mappings between visible and thermal images \cite{ref11,ref12}, and have also been introduced into  UAV thermal geo-localization \cite{ref4}. However, adversarial training and image generation usually involve multiple competing optimization objectives. They may introduce training instability and weaken structural information required for fine-grained matching.

Normalization-based methods provide another way to correct modality differences. AdaIN transfers style statistics by adjusting the channel-wise means and variances of feature maps \cite{ref29}, while BIN learns an adaptive combination of batch normalization and instance normalization to obtain more domain-invariant representations \cite{ref30}. These studies show that part of the modality discrepancy can be alleviated through lightweight adjustment of feature statistics. Nevertheless, correcting global statistics alone does not guarantee the preservation of spatial structures shared by heterogeneous images.

Recent thermal-to-visible cross-modal retrieval studies have explored heterogeneous image representation learning from several complementary perspectives, including multi-level feature alignment \cite{ref31,ref32}, modality-shared and modality-specific representation learning \cite{ref33,ref34}, multi-scale semantic correlation modeling \cite{ref35}, intermediate-modality interaction \cite{ref36,ref37}, and cross-modal representation distillation \cite{ref38}. These methods reduce modality discrepancies and improve feature discriminability to some extent, but their explicit modeling of cross-modal shared structures remains limited. In multimodal remote sensing, studies on optical--SAR fusion and building extraction have also shown that simple feature concatenation or statistical alignment may disrupt spatial relationships between images acquired by different sensors, whereas preserving structural and geometric consistency improves the reliability of cross-modal representations \cite{ref15,ref16}. Consequently, for matching thermal UAV images with satellite imagery, merely reducing modality distribution discrepancies is insufficient. It is imperative to identify scene structures that remain invariant across these modalities and leverage them as robust priors for cross-modal matching.

\subsection{Spatial Structure Feature Learning}

Directional patterns, boundaries, and large-scale spatial layouts constitute important structural information in remote-sensing images. These features are generally less sensitive to sensor type and imaging conditions and can therefore provide relatively stable cues for heterogeneous image analysis and matching. Classical oriented filtering methods, such as the Kirsch operator, extract gradient responses along multiple discrete directions and describe local structural orientations without relying on absolute intensity values \cite{ref39}. In deep learning, multi-level and multi-scale feature fusion has been used to enhance edge representations \cite{ref40}, deep Hough transforms have been introduced to explicitly model semantic line structures \cite{ref41}, and edge-attention mechanisms have been used to enhance shared structural information between infrared and visible images \cite{ref42}. These studies indicate that edges, contours, and directional patterns can serve as more stable intermediate representations than raw appearance information.

In remote-sensing scenes, variations in object orientation and spatial organization have also motivated orientation-sensitive and rotation-aware feature extraction methods \cite{ref43}. In addition, conventional convolutional networks represented by ResNet \cite{ref45} and hierarchical vision Transformers represented by Swin Transformer \cite{ref44} have been widely adopted for visual representation learning. These architectures have also been applied to cross-view UAV--satellite geo-localization and remote-sensing image analysis \cite{ref46,ref47}. Hierarchical features and cross-region context modeling help capture both local boundaries and large-scale spatial layouts. However, existing studies generally focus on edge detection, directional modeling, rotation-aware object recognition, or generic visual feature extraction. They do not explicitly enhance directional structures and low-frequency spatial organization for cross-modal matching between thermal UAV images and satellite images, nor do they further address the modality discrepancy remaining in the structure-enhanced features.


\section{METHODOLOGY}
\label{sec:methodology}
This section first describes the UAV geo-localization task based on cross-view matching between thermal and satellite images, and then introduces the proposed method.

\subsection{Task Definition}
Let the thermal UAV query image be denoted as \(I^t\), and the satellite database image be denoted as \(I^s\). The objective of UAV geo-localization based on cross-view matching between thermal and satellite images is to retrieve the satellite image corresponding to the geographic location of a given thermal query image \(I^t\) from a large-scale geo-tagged satellite reference database \(\mathcal{D}=\{I_i^s\}_{i=1}^{N}\). Specifically, the geo-localization model maps the query image and satellite images into global retrieval descriptors, and then performs nearest-neighbor retrieval according to descriptor similarity. This process can be formulated as
\begin{equation}
\hat{i}
=
\mathop{\arg\max}_{i \in \{1,\ldots,N\}}
s\left(z^t,z_i^s\right),
\label{eq:retrieval}
\end{equation}
where \(z^t\) denotes the global descriptor of the query image \(I^t\), \(z_i^s\) denotes the global descriptor of the \(i\)-th satellite database image \(I_i^s\), \(s(\cdot,\cdot)\) denotes the similarity function between descriptors, and \(\hat{i}\) denotes the index of the retrieved satellite image.

\subsection{Overall Framework}
\begin{figure*}[!t] 
\centering
\includegraphics[width=\textwidth]{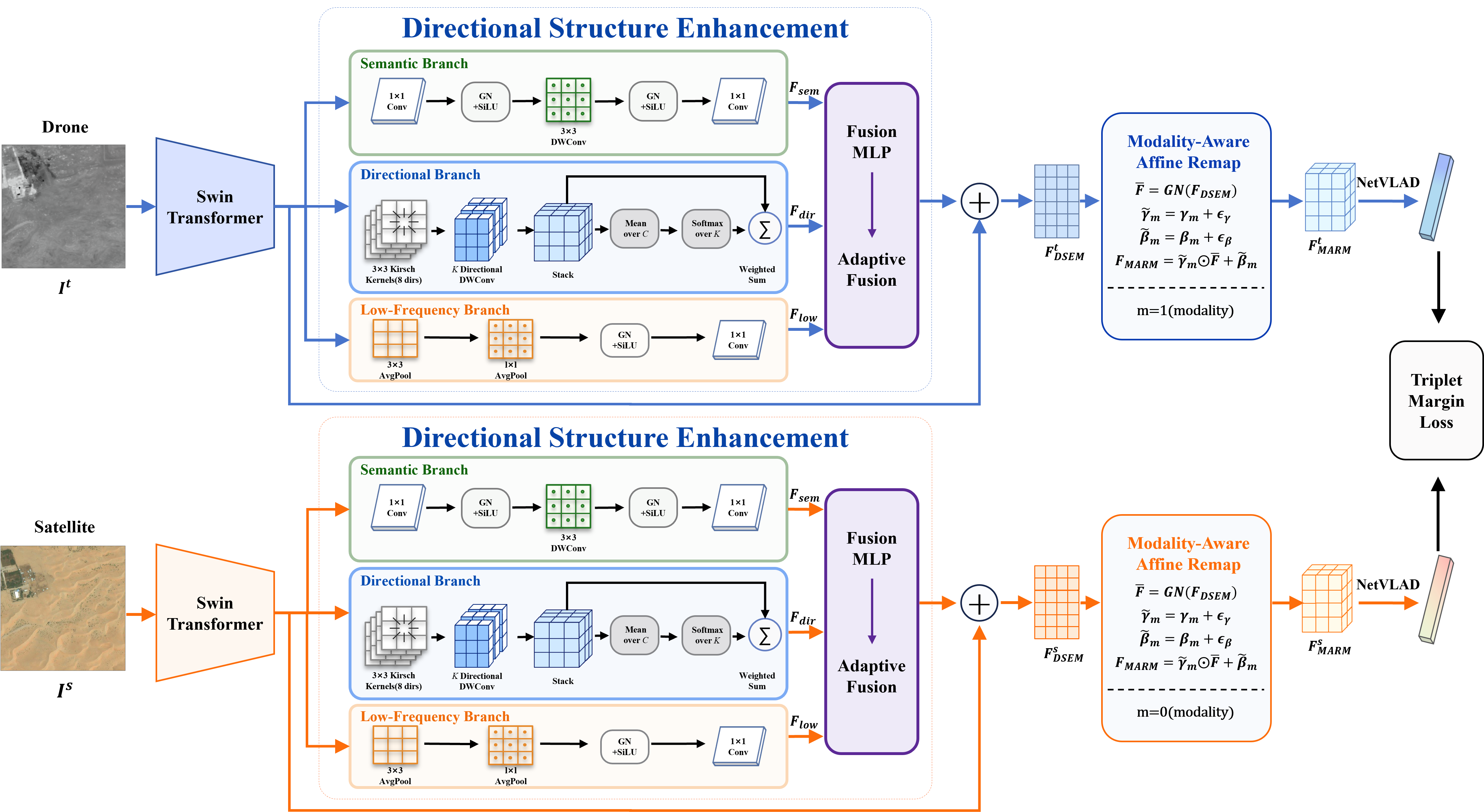}
\caption{The pipline of the proposed structure-aware matching framework for UAV thermal-to-satellite geo-localization. The thermal query image and  satellite image are fed into a shared-parameter Swin Transformer backbone to extract feature maps. The backbone features are then enhanced by the Directional Structure Enhancement Module, where adaptive fusion of the semantic-preserving branch, multi-directional response branch, and low-frequency structure branch strengthens directional structures and low-frequency spatial organizations that are more stable under cross-modal conditions. Subsequently, the Modality-Aware Affine Remapping module performs lightweight channel-wise statistical correction on the structure-enhanced features, reducing the statistical distribution shift between thermal and satellite features.}
\label{fig:framework} 
\end{figure*}

To address the significant modality discrepancy between thermal and satellite images, as well as the lack of strong localization anchors in low-semantic scenes such as deserts and bare lands, we propose a cross-modal UAV thermal geo-localization framework centered on structure-prioritized representation learning. Specifically, the framework focuses on learning relatively stable information, including surface directional patterns, region boundaries, and low-frequency spatial layouts, which is less sensitive to modality-specific texture and semantic differences and is therefore more suitable for cross-modal localization. The overall pipeline is illustrated in Fig.~\ref{fig:framework}.

The query image \(I^t\) and the database image \(I^s\) are first fed into a shared-parameter Swin Transformer backbone to extract deep feature maps \(F^t\) and \(F^s\), respectively. The DSEM module then performs structure-prioritized enhancement on the backbone features, explicitly strengthening structural cues that remain relatively stable across thermal and satellite modalities, such as directional boundaries and low-frequency terrain trends. This produces the structure-enhanced features \(F_{\mathrm{DSEM}}^{t}\) and \(F_{\mathrm{DSEM}}^{s}\). Subsequently, the MARM module conducts lightweight channel-wise statistical correction on the structure-enhanced features according to the modality labels, reducing the residual distribution shift between thermal and satellite features and generating the aligned features \(F_{\mathrm{MARM}}^{t}\) and \(F_{\mathrm{MARM}}^{s}\). In this pipeline, DSEM first addresses the question of which features are more stable across modalities, while MARM further addresses how these features can be aligned into a unified distribution. The two modules form a synergistic feature processing pipeline. Finally, the aligned features are compressed along the channel dimension and aggregated  into global retrieval descriptors. During training, the model is optimized with a triplet loss, which pulls the features of matched image pairs closer while pushing the features of unmatched image pairs apart. 

The following subsections describe the proposed DSEM and MARM modules in detail.

\subsection{Directional Structure Enhancement Module}
Between thermal UAV images and satellite images, the same geographic location often exhibits substantially different color, brightness, and local texture responses due to the distinct imaging mechanisms. This discrepancy is particularly pronounced in low-semantic scenes such as deserts and bare lands, where images usually lack strong localization anchors such as dense buildings or clear road intersections. Nevertheless, thermal and satellite images still maintain relatively stable spatial structural consistency in terms of surface stripe orientations, weak boundary extension trends, and coarse-grained regional layouts, which can be seen from Fig.~\ref{fig:boson_examples}. Motivated by this observation, we propose the DSEM module, which performs structure-prioritized enhancement on deep features generated by the backbone. As illustrated in the framework of Fig.~\ref{fig:framework}, DSEM consists of three parallel branches: a semantic-preserving branch, a directional response branch, and a low-frequency structure branch. These branches complement the backbone features from different perspectives and are then integrated through adaptive fusion and residual connection to obtain the final output.

\subsubsection{Semantic-Preserving Branch}
The first branch of DSEM is the semantic-preserving branch, which aims to maintain the contextual information and scene-level discriminative capability already encoded in deep features, preventing semantic degradation caused by excessive emphasis on structural responses. The semantics does not refer to category-level semantics in conventional classification tasks, but rather to the regional organization, contextual dependencies, and high-level representations of the overall scene layout that are implicitly embedded in the deep backbone features.

In implementation, the semantic-preserving branch is modeled by a lightweight convolutional block, consisting of a \(1 \times 1\) convolution, GroupNorm, SiLU, a \(3 \times 3\) depthwise convolution, and another \(1 \times 1\) convolution. This design preserves inter-channel interactions while supplementing local spatial modeling capability at a low computational cost. The output of this branch is denoted as \(F_{\rm sem}\).

\subsubsection{Directional Response Branch}
The directional response branch is the core component of DSEM and is designed to explicitly model the dominant orientation and orientation-sensitive responses of local structures. Different from global modeling strategies that assume a single dominant orientation for the entire image, we argue that in UAV thermal-to-satellite cross-modal localization, different spatial locations may correspond to different local structural trends. Therefore, it is necessary to adaptively identify the more salient directional patterns at the pixel-location level.

To this end, we construct eight direction-sensitive units for the input feature, each implemented by a \(3 \times 3\) depthwise convolution and initialized with a Kirsch-style directional template. Let the \(k\)-th direction-sensitive convolutional unit be denoted as \(D_k(\cdot)\). The response of the \(k\)-th direction can be written as
\begin{equation}
R_k = D_k(F), \quad k=1,2,\ldots,8 .
\label{eq:direction_response}
\end{equation}

To measure the preference of each spatial location for different directions, we first average each directional response along the channel dimension to obtain a location-dependent directional score map:
\begin{equation}
s_k = \operatorname{Mean}_{c}(R_k),
\label{eq:direction_score}
\end{equation}
where \(s_k \in \mathbb{R}^{H \times W}\). Then, a softmax operation is performed along the directional dimension to obtain the normalized weight of each direction at each spatial location:
\begin{equation}
\alpha_k(i,j)
=
\frac{\exp \left(s_k(i,j)\right)}
{\sum_{\ell=1}^{8} \exp \left(s_{\ell}(i,j)\right)} .
\label{eq:direction_weight}
\end{equation}

Finally, the eight directional responses are adaptively fused using these directional weights to obtain the direction-enhanced structural feature:
\begin{equation}
F_{\mathrm{dir}}
=
\sum_{k=1}^{8} \alpha_k \odot R_k ,
\label{eq:direction_fusion}
\end{equation}
where \(\odot\) denotes element-wise multiplication and \(\alpha_k\) is broadcast from the spatial locations to the channel dimension.

The physical meaning of this branch can be understood as follows: for each spatial location in the feature map, eight directional probes are used to test which type of directional pattern its local structure is closer to, and the responses are then fused according to adaptive weights. Unlike conventional convolutions that mix directional information into general local texture responses, we adopt an explicit multi-directional response modeling strategy to separately perceive and fuse structural patterns in different directions, which is therefore more beneficial for preserving the stable spatial directional cues existing in cross-modal matching.

\subsubsection{Low-Frequency Structure Branch}
Although the directional response branch can strengthen local structural orientations, in desert, bare-land, and weak-road scenes, cross-modal stable information is not always manifested as clear edges or local directions. Instead, it is often reflected in large-scale, slowly varying terrain trends and coarse-grained regional layouts. To supplement this type of information, we design a low-frequency structure branch, which suppresses fragmented high-frequency interference through local smoothing and larger-receptive-field convolution, while enhancing the response to  low-frequency large-scale structural patterns.

This branch first smooths local details through average pooling, and then uses a large-kernel depthwise convolution to model slowly varying structures under a larger receptive field, followed by GroupNorm, SiLU, and a \(1 \times 1\) convolution for feature mapping. In our experiments, the convolution kernel size of the low-frequency branch is usually set to 5 and is validated among several candidate scales. The output of this branch is denoted as \(F_{\rm low}\).

\subsubsection{Adaptive Fusion}
Since the three branches focus on different types of information, DSEM does not adopt simple summation for fusion. Instead, it adaptively predicts the contribution of each branch according to the global context of the input feature. Specifically, global average pooling is first applied to the input feature \(F\) to extract channel-level global context, which is then fed into a lightweight MLP followed by a softmax operation to obtain the normalized fusion weights of the semantic-preserving branch, directional response branch, and low-frequency structure branch:
\begin{equation}
[\omega_{\rm sem}, \omega_{\rm dir}, \omega_{\rm low}]
=
\operatorname{Softmax}(\operatorname{MLP}(\operatorname{GAP}(F))).
\end{equation}
where \(\omega_{\rm sem}\), \(\omega_{\rm dir}\), and \(\omega_{\rm low}\) denote the weights of the semantic-preserving branch, directional response branch, and low-frequency structure branch, respectively, and satisfy
\begin{equation}
\omega_{\rm sem}+\omega_{\rm dir}+\omega_{\rm low}=1.
\end{equation}

Based on these adaptive weights, the three branch outputs are fused as
\begin{equation}
F_{\rm fuse} = \omega_{\rm sem}F_{\rm sem} + \omega_{\rm dir}F_{\rm dir} + \omega_{\rm low}F_{\rm low},
\end{equation}
where the weights are broadcast to match the spatial and channel dimensions of the branch features. This adaptive fusion strategy enables the model to balance different structural cues according to the scene content. For example, when local directional structures are obvious, the directional response branch can obtain a higher weight; when the scene lacks clear edges but has a stable regional layout, the low-frequency structure branch may contribute more.

The final output of DSEM is obtained by applying normalization after the residual connection:
\begin{equation}
F_{\rm DSEM} = \operatorname{GN}(F + F_{\rm fuse}).
\end{equation}

\subsection{Modality-Aware Affine Remapping Module}
Although DSEM has pushed the features toward a more stable structural representation space, there are still obvious differences in the feature statistical distributions between the thermal query and the satellite images. For example, the two modalities are usually inconsistent in channel mean, response magnitude, and the activation strength of certain structural channels. If such structure-enhanced features are directly fed into a unified global aggregator, the subsequent global descriptor generation process can still be affected by modality shifts, thereby limiting cross-modal retrieval performance. Since this shift is mainly reflected at the channel-wise statistical level rather than the spatial geometric level, we do not adopt complex cross-modal interactions or spatial transformations. Instead, we propose a lightweight MARM to perform statistical correction on features from different modalities without destroying the spatial structures of these features.

For the output feature \(F_{\mathrm{DSEM}}\) of DSEM, GroupNorm is first applied to obtain the normalized feature:
\begin{equation}
\bar{F} = \mathrm{GN}\left(F_{\mathrm{DSEM}}\right).
\label{eq:marm_norm}
\end{equation}

Then, according to the input modality label \(m\), the corresponding channel-wise affine parameters \(\gamma_m\) and \(\beta_m\) are selected to perform modality-aware remapping:
\begin{equation}
F_{\mathrm{MARM}}
=
\gamma_m \odot \bar{F} + \beta_m ,
\label{eq:marm_remap}
\end{equation}
where \(m=0\) denotes the satellite image and \(m=1\) denotes the thermal UAV image. \(\gamma_m, \beta_m \in \mathbb{R}^{C}\) are learnable parameters corresponding to modality, and are independently learned to provide targeted correction for the channel response distributions of the two modalities.

To further improve model robustness, we introduce lightweight Gaussian perturbations to the affine parameters during training. Specifically, small random noises are injected into \(\gamma_m\) and \(\beta_m\) during affine remapping, which can be formulated as
\begin{equation}
\tilde{\gamma}_m = \gamma_m + \epsilon_{\gamma},
\quad
\tilde{\beta}_m = \beta_m + \epsilon_{\beta},
\label{eq:marm_perturbation}
\end{equation}
where
\begin{equation}
\epsilon_{\gamma} \sim \mathcal{N}\left(0,\sigma_{\gamma}^{2}\right),
\quad
\epsilon_{\beta} \sim \mathcal{N}\left(0,\sigma_{\beta}^{2}\right).
\label{eq:marm_noise}
\end{equation}

Thus, the remapping process during training is written as
\begin{equation}
F_{\mathrm{MARM}}
=
\tilde{\gamma}_m \odot \bar{F} + \tilde{\beta}_m .
\label{eq:marm_training}
\end{equation}

This lightweight perturbation can be regarded as imposing a random regularization constraint on the modality-aware affine correction process, which helps suppress overfitting of the parameters to sample-specific statistical patterns in the training set and enhances the generalization capability in testing scenarios. In our experiments, \(\sigma_{\gamma}=0.02\) and \(\sigma_{\beta}=0.01\).

\section{Experiments}
\label{sec:experiments}

In this section, we demonstrate the superiority of our method through experiments. The experimental setup is first outlined, including datasets, metrics, implementation details, and comparison methods. We then report quantitative results on the Boson-nighttime and Thermal-UAV-Paired datasets. After that, we provide an in-depth analysis of the DSEM and MARM modules, supported by ablation studies, parameter sensitivity tests, and visualizations.

\subsection{Experimental Setup}
\subsubsection{Dataset}
We conduct experiments on the Boson-nighttime dataset and the newly constructed Thermal-UAV-Paired dataset. 

The Boson-nighttime dataset was proposed by Xiao et al. \cite{ref4} for cross-modal geo-localization between thermal UAV images and satellite images. The dataset was collected using a FLIR Boson thermal camera with a sensor focal length of \(8.7~\mathrm{mm}\), a resolution of 640p, and a horizontal field of view of \(50^{\circ}\). The captured images are approximately nadir-view images, with a spatial resolution of about \(1~\mathrm{m/px}\). Data collection involved six nighttime flights, conducted from 9:00 PM to 4:00 AM the next day.  The entire dataset covers a ground area of approximately \(33~\mathrm{km}^{2}\). The scenes are dominated by desert areas, with a small number of farmlands, roads, and building regions. To construct paired samples for cross-modal localization, the original thermal multi-view images are first used to reconstruct a thermal map through structure-from-motion (SfM). The reconstructed map is then orthorectified and spatially aligned with satellite imagery of the same area, thereby obtaining paired thermal-satellite images. The satellite reference images of the corresponding area are obtained from Bing satellite maps. To construct paired samples, the thermal map is cropped into \(512 \times 512\)-pixel image patches with a sliding step of 35 pixels, and each patch is paired with the satellite image patch at the corresponding spatial location. The dataset uses the coverage areas of three out of the six flights for training and validation, while the coverage areas of the remaining three flights are used for testing. The train/validation/test splits contain \(10256\), \(13011\), and \(26568\) pairs of satellite-thermal image patches, respectively.

\begin{figure}[!t] 
\centering
\includegraphics[width=\linewidth]{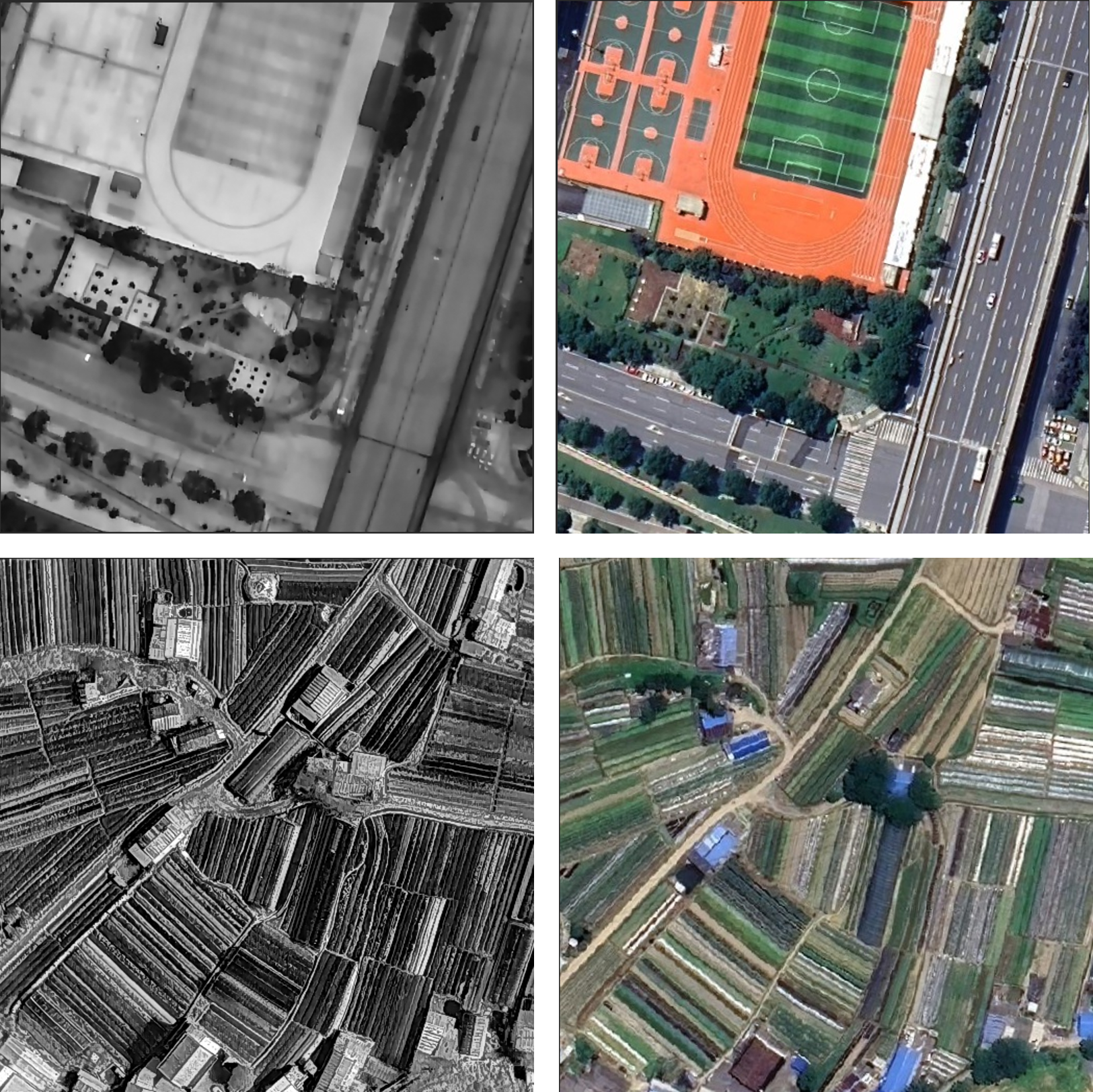}
\caption{Typical scene examples from the Thermal-UAV-Paired dataset. The left column shows  UAV thermal images, and the right column shows satellite reference images cropped according to the corresponding center locations. Unlike the desert scenes in Boson-nighttime dataset, this dataset contains more complex ground object types, such as sports fields, roads, and farmlands.}
\label{fig:thermal_examples} 
\end{figure}

To rigorously evaluate our proposed method, we have also designed the Thermal-UAV-Paired dataset as shown in Fig.~\ref{fig:thermal_examples}, which is derived from the Thermal-UAV dataset \cite{ref17} by establishing precise one-to-one correspondences between thermal UAV images and satellite imagery. The Thermal-UAV dataset contains thermal UAV images and large-scale satellite orthophotos of the corresponding areas.  We crop the corresponding satellite image patch from the satellite orthophoto according to the center location of each thermal UAV image, and establish a one-to-one paired relationship between thermal UAV images and satellite images, making it suitable for image-level cross-modal retrieval evaluation. Unlike Boson-nighttime dataset, which constructs paired samples by sliding-window cropping on a thermal orthophoto map, Thermal-UAV-Paired directly uses the original thermal UAV  images as query images and generates the corresponding satellite reference image patches according to their center locations. The constructed Thermal-UAV-Paired benchmark contains training, validation, and testing sets, including 7886, 1407, and 2350 thermal-satellite image pairs, respectively. Each sample retains both the center latitude-longitude coordinates and projected coordinates. Table~\ref{tab:dataset_statistics} summarizes the numbers of paired samples in the training, validation, and testing splits of the two datasets.

\begin{table}[!t]
	\caption{Statistics of the Boson-nighttime and Thermal-UAV-Paired datasets.}
	\label{tab:dataset_statistics}
	\centering
	\footnotesize
	\renewcommand{\arraystretch}{1.12}
	\begin{tabular*}{0.82\columnwidth}{@{\hspace{4pt}\extracolsep{\fill}}lcc@{\hspace{4pt}}}
		\toprule
		Dataset & Split & Paired Samples \\
		\midrule
		\multirow{3}{*}{Boson-nighttime}
		& Training   & 10256 \\
		& Validation & 13011 \\
		& Testing    & 26568 \\
		\midrule
		\multirow{3}{*}{Thermal-UAV-Paired}
		& Training   & 7886 \\
		& Validation & 1407 \\
		& Testing    & 2350 \\
		\bottomrule
	\end{tabular*}
\end{table}

\subsubsection{Evaluation Metrics}
We adopt the following metrics to evaluate the performance of thermal-to-satellite cross-modal geo-localization.

\emph{Recall@K} \((\mathrm{R}\text{@}K)\): This metric measures whether the geographically corresponding satellite image of each thermal query image appears in the top-\(K\) retrieved candidates. The criterion is as follows: if the geographic distance between the retrieved satellite image and the query image is within \(50~\mathrm{m}\), it is regarded as a correct match. \(K \in \{1,5\}\). \(\mathrm{R}\text{@}1\) reflects the precise matching capability of the model, while \(\mathrm{R}\text{@}5\) reflects the recall capability of the model within a small candidate set. Higher values indicate better performance.

\emph{Recall@K with prior location threshold} \(d\) \((\mathrm{R}_{d}\text{@}K)\): This metric adds a search range constraint based on \(\mathrm{R}\text{@}K\), limiting the retrieval range to a region centered at the query location with a radius of \(d\) meters. This metric simulates scenarios in which a rough location is known during actual flight, such as the localization result of the previous frame or an inertial estimate, allowing the model to perform precise retrieval within a limited candidate range. We report the results with \(d=512~\mathrm{m}\), denoted as \(\mathrm{R}_{512}\text{@}1\) and \(\mathrm{R}_{512}\text{@}5\). Higher values indicate better performance.

\emph{L2 distance error with prior location threshold} \(d\) \((L_{d}^{2})\): Within the search radius of \(d\) meters, this metric takes the Euclidean distance between the geographic location of the Top-1 retrieved result and the ground-truth query location as the localization error, measured in meters. This metric evaluates the absolute positioning accuracy of the model within a coarse localization range. We report the results with \(d=512~\mathrm{m}\), denoted as \(L_{512}^{2}~(\mathrm{m})\). Lower values indicate better performance.

\subsection{Comparative Experiments}
Tables~\ref{tab:comparison_boson} and ~\ref{tab:comparison_thermal} present the comparison results on the Boson-nighttime and Thermal-UAV-Paired test sets, respectively. To comprehensively evaluate the performance of the proposed method in matching thermal UAV images with satellite images, we consider three categories of comparison methods. The first category consists of general backbone baselines, including ResNet-18, ResNet-50, DINOv3-ConvNeXt-T/B, DINOv3-ViT-S, and Swin-T. All these backbone networks adopt the same NetVLAD feature aggregation strategy without introducing additional cross-modal processing modules, and are used to compare the performance differences of different feature extraction networks under a unified retrieval framework. The second category consists of  UAV thermal geo-localization methods, including SGM w/o Gen. and SGM \cite{ref4}. These methods are specifically designed for nighttime thermal UAV localization, and alleviate the modality discrepancy between thermal UAV images and satellite images through domain-adversarial learning and GAN-based generative sample expansion. The third category consists of thermal-visible cross-modal retrieval methods, including MCJA \cite{ref34}, IDKL \cite{ref35}, MSCMNet \cite{ref36}, IFALNet \cite{ref37}, and AnyThermal \cite{ref38}. These methods cover typical cross-modal processing strategies such as feature alignment, modality-shared representation, multi-scale semantic modeling, intermediate modality interaction, and general thermal representation learning. Since the task studied in this paper also involves heterogeneous feature matching between thermal images and RGB images, we adapt these methods to the matching task between thermal UAV images and RGB satellite images as supplementary comparison methods.

On the Boson-nighttime dataset, the proposed method achieves the best results on four metrics, including \(\mathrm{R}\text{@}1\), \(\mathrm{R}\text{@}5\), \(\mathrm{R}_{512}\text{@}1\), and \(\mathrm{R}_{512}\text{@}5\). Among the general backbone baselines, ResNet-18 and ResNet-50 obtain \(\mathrm{R}\text{@}1\) values of \(63.2\%\) and \(63.5\%\), respectively, indicating that traditional CNN features have difficulty establishing stable cross-modal correspondences in nighttime thermal scenes with weak textures. DINOv3-ViT-S and Swin-T improve \(\mathrm{R}\text{@}1\) to \(83.2\%\) and \(84.1\%\), respectively, showing that stronger visual representations and hierarchical feature structures can improve the matching between thermal images and satellite images. However, these baselines still rely only on general image features and do not explicitly handle the structural differences and modality shift between thermal and satellite images. In contrast, SAMF improves \(\mathrm{R}\text{@}1\) to \(93.6\%\) based on the same Swin-T backbone, demonstrating that DSEM enhances directional structures and low-frequency spatial layouts, and MARM corrects modality discrepancies, thereby further improving cross-modal retrieval performance beyond general backbone features.

Compared with  UAV thermal geo-localization methods, SGM w/o Gen. obtains an \(\mathrm{R}\text{@}1\) of \(72.1\%\) and an \(L_{512}^{2}\) of \(52.2~\mathrm{m}\) on Boson-nighttime dataset, indicating that cross-modal alignment based on domain-adversarial learning still has difficulty fully adapting to weak-texture nighttime scenes when only the original paired data are used. By introducing GAN-based generative augmented samples, the complete SGM improves \(\mathrm{R}\text{@}1\) to \(92.1\%\) and reduces \(L_{512}^{2}\) to \(14.7~\mathrm{m}\), showing that generative data expansion can effectively alleviate the problem of insufficient training samples. However, without using generative augmented samples, the proposed method still achieves \(\mathrm{R}\text{@}1 = 93.6\%\), \(\mathrm{R}\text{@}5 = 97.4\%\), \(\mathrm{R}_{512}\text{@}1 = 98.2\%\), and \(\mathrm{R}_{512}\text{@}5 = 99.6\%\), all higher than the corresponding recall results of the complete SGM. This indicates that, compared with indirectly expanding the training distribution through image generation, directly enhancing cross-modally stable structures at the feature level and performing modality correction is more suitable for thermal UAV localization in weak-semantic scenes such as deserts, bare lands, and sparse vegetation in Boson-nighttime dataset.

The thermal-visible cross-modal retrieval methods in Table~\ref{tab:comparison_boson} further show the performance of different cross-modal representation learning strategies on the Boson-nighttime dataset. MCJA, IDKL, and IFALNet reduce modality discrepancies from the perspectives of cross-layer feature alignment, modality-shared representation learning, and intermediate modality interaction, respectively. Their \(\mathrm{R}\text{@}5\) and \(\mathrm{R}_{512}\text{@}5\) results reach relatively high levels, indicating that these methods can retain more correct matches among the candidate results. However, their \(\mathrm{R}\text{@}1\) values are \(61.7\%\), \(56.2\%\), and \(64.7\%\), respectively, which are significantly lower than \(93.6\%\) of the proposed method. This suggests that relying only on feature alignment or modality-shared representation is still insufficient to ensure that the correct sample ranks first in global retrieval. MSCMNet introduces multi-scale semantic correlation modeling and achieves favorable results on \(\mathrm{R}_{512}\text{@}1\), \(\mathrm{R}_{512}\text{@}5\), and \(L_{512}^{2}\), indicating that multi-scale features help improve the matching accuracy within a local candidate range. However, its \(\mathrm{R}\text{@}1\) is still \(18.1\) percentage points lower than that of the proposed method. AnyThermal learns general thermal epresentations through cross-modal distillation, but its \(\mathrm{R}\text{@}1\) and \(\mathrm{R}\text{@}5\) are only \(52.18\%\) and \(70.01\%\), respectively, indicating that directly applying general thermal representations to nighttime thermal UAV-satellite image matching remains insufficient. In contrast, beyond correcting modality discrepancies, SAMF further explicitly enhances stable ground cues such as directional structures, region boundaries, and low-frequency spatial layouts, thereby achieving higher global recall on Boson-nighttime dataset.

On the Thermal-UAV-Paired dataset, the proposed method also achieves the best results, with \(\mathrm{R}\text{@}1\), \(\mathrm{R}\text{@}5\), \(\mathrm{R}_{512}\text{@}1\), \(\mathrm{R}_{512}\text{@}5\), and \(L_{512}^{2}\) reaching \(97.0\%\), \(99.4\%\), \(98.1\%\), \(99.8\%\), and \(4.8~\mathrm{m}\), respectively. Unlike Boson-nighttime dataset, which mainly consists of weak-semantic scenes such as deserts, bare lands, and sparse vegetation, Thermal-UAV-Paired contains more artificial ground objects and regional structures, including roads, sports fields, and farmlands. Meanwhile, this dataset is composed of thermal UAV images and satellite image patches cropped according to their corresponding locations, and does not contain additional unpaired satellite images for generative data expansion. Therefore, SGM in Table~\ref{tab:comparison_thermal} is evaluated under the setting without generative augmented samples. Most backbone baselines have already achieved high recall rates on this dataset. For example, DINOv3-ViT-S and DINOv3-ConvNeXt-T achieve \(\mathrm{R}\text{@}1\) values of \(94.6\%\) and \(94.1\%\), respectively, indicating that the clearer road boundaries, field contours, and farmland textures in this dataset provide more usable cues for cross-modal matching. On this basis, SAMF still further achieves the highest recall rates and the lowest localization error, demonstrating that the proposed directional structure enhancement and modality-aware correction are not only suitable for weak-texture nighttime scenes in Boson-nighttime dataset, but can also generalize to thermal UAV scenes with more artificial structures.

\begin{table*}[!t]
\caption{Comparison with existing methods on the Boson-nighttime test set. The best results are highlighted in \textbf{bold}.}
\label{tab:comparison_boson}
\centering
\footnotesize
\setlength{\tabcolsep}{3.8pt}
\renewcommand{\arraystretch}{1.12}
\begin{tabular*}{0.98\textwidth}{@{\hspace{4pt}}l@{\extracolsep{\fill}}cccccc@{\hspace{4pt}}}
\toprule
Method & Venue & $\mathrm{R}@1$ & $\mathrm{R}@5$ & $R_{512}@1$ & $R_{512}@5$ & $L_2^{512}$ (m) \\
\midrule
\multicolumn{7}{l}{\textit{Backbone baselines}} \\
ResNet-18          & -- & 63.2 & 75.6 & 80.0 & 89.5 & 24.6 \\
ResNet-50          & -- & 63.5 & 75.9 & 78.1 & 88.5 & 26.4 \\
DINOv3-ConvNeXt-T  & -- & 65.7 & 76.6 & 82.8 & 90.8 & 35.9 \\
DINOv3-ConvNeXt-B  & -- & 72.9 & 82.4 & 86.7 & 94.1 & 36.6 \\
DINOv3-ViT-S       & -- & 83.2 & 90.0 & 93.9 & 96.9 & 19.4 \\
Swin-T             & -- & 84.1 & 90.9 & 93.7 & 97.7 & 26.5 \\
\midrule
\multicolumn{7}{l}{\textit{Geo-localization Methods}} \\
SGM w/o Gen.~\cite{ref4}   & CVPR'23          & 72.1  & 83.9  & 84.5 & 93.7  & 52.2 \\
SGM~\cite{ref4}            & CVPR'23          & 92.1  & 96.9  & 96.5 & 99.1  & 14.7 \\
MCJA~\cite{ref34}          & TCSVT'24         & 61.7  & 92.4  & 95.8 & 98.9  & 15.5 \\
IDKL~\cite{ref35}          & CVPR'24          & 56.2  & 91.1  & 95.4 & 98.4  & 18.4 \\
MSCMNet~\cite{ref36}       & PR'25            & 75.5  & 96.8  & 97.9 & 99.5  & \textbf{6.7} \\
IFALNet~\cite{ref37}       & Neurocomputing'26 & 64.7  & 96.2  & 97.2 & 99.3  & 8.1 \\
AnyThermal~\cite{ref38}    & ICRA'26          & 52.18 & 70.01 & 80.3 & 90.14 & 72.6 \\
\midrule
SAMF               & Ours & \textbf{93.6} & \textbf{97.4} & \textbf{98.2} & \textbf{99.6} & 12.3 \\
\bottomrule
\end{tabular*}
\end{table*}

\begin{table*}[!t]
\caption{Comparison with existing methods on the Thermal-UAV-Paired test set. The best results are highlighted in \textbf{bold}.}
\label{tab:comparison_thermal}
\centering
\footnotesize
\setlength{\tabcolsep}{3.8pt}
\renewcommand{\arraystretch}{1.12}
\begin{tabular*}{0.98\textwidth}{@{\hspace{4pt}}l@{\extracolsep{\fill}}cccccc@{\hspace{4pt}}}
\toprule
Method & Venue & $\mathrm{R}@1$ & $\mathrm{R}@5$ & $R_{512}@1$ & $R_{512}@5$ & $L_2^{512}$ (m) \\
\midrule
\multicolumn{7}{l}{\textit{Backbone baselines}} \\
ResNet-18          & -- & 93.1 & 98.0 & 94.9 & 99.0 & 12.8 \\
ResNet-50          & -- & 93.3 & 97.9 & 94.6 & 98.6 & 13.1 \\
DINOv3-ConvNeXt-T  & -- & 94.1 & 98.3 & 96.7 & 99.1 & 8.5 \\
DINOv3-ConvNeXt-B  & -- & 77.7 & 88.0 & 84.7 & 94.3 & 41.5 \\
DINOv3-ViT-S       & -- & 94.6 & 98.1 & 97.5 & 99.5 & 6.3 \\
Swin-T             & -- & 92.8 & 98.5 & 94.6 & 99.0 & 12.3 \\
\midrule
\multicolumn{7}{l}{\textit{Geo-localization Methods}} \\
SGM w/o Gen.~\cite{ref4}   & CVPR'23          & 83.9 & 93.3 & 87.2 & 95.9 & 30.1 \\
MCJA~\cite{ref34}          & TCSVT'24         & 82.8 & 95.9 & 95.7 & 98.2 & 11.6 \\
IDKL~\cite{ref35}          & CVPR'24          & 85.2 & 97.7 & 97.3 & 99.5 & 6.2 \\
MSCMNet~\cite{ref36}       & PR'25            & 87.7 & 96.5 & 96.4 & 97.9 & 9.7 \\
IFALNet~\cite{ref37}       & Neurocomputing'26 & 84.5 & 96.9 & 96.7 & 98.9 & 7.8 \\
AnyThermal~\cite{ref38}    & ICRA'26          & 80.5 & 97.5 & 81.0 & 98.3 & 5.0 \\
\midrule
SAMF               & Ours & \textbf{97.0} & \textbf{99.4} & \textbf{98.1} & \textbf{99.8} & \textbf{4.8} \\
\bottomrule
\end{tabular*}
\end{table*}

\subsection{Ablation Experiments}
To further analyze the contribution of each module to the final performance, we conduct step-by-step ablation experiments on DSEM and MARM modules under the same backbone network and training settings. The results are shown in Table~\ref{tab:ablation_modules}.

\begin{table*}[!t]
\caption{Ablation results of DSEM and MARM modules.
The best results are highlighted in \textbf{bold}.}
\label{tab:ablation_modules}
\centering
\renewcommand{\arraystretch}{1.12}
\begin{tabular*}{0.82\textwidth}{@{\hspace{4pt}}l@{\extracolsep{\fill}}ccccccc@{\hspace{4pt}}}
\toprule
Backbone 
& DSEM 
& MARM 
& \(\mathrm{R}\text{@}1\) 
& \(\mathrm{R}\text{@}5\) 
& \(\mathrm{R}_{512}\text{@}1\) 
& \(\mathrm{R}_{512}\text{@}5\) 
& \(L_{512}^{2}~(\mathrm{m})\) \\
\midrule
Swin-T &  &  & 84.1 & 90.9 & 93.7 & 97.7 & 26.5 \\
Swin-T & \(\checkmark\) &  & 82.8 & 91.7 & 93.4 & 98.2 & 25.5 \\
Swin-T &  & \(\checkmark\) & 83.3 & 91.1 & 93.3 & 97.7 & 29.4 \\
Swin-T & \(\checkmark\) & \(\checkmark\) & \textbf{93.6} & \textbf{97.4} & \textbf{98.2} & \textbf{99.6} & \textbf{12.3} \\
\bottomrule
\end{tabular*}
\end{table*}

Table~\ref{tab:ablation_modules} presents the ablation results of Swin-T baseline, Swin-T+DSEM, Swin-T+MARM, and Swin-T+DSEM+MARM. It can be seen that when DSEM or MARM is introduced alone, the model performance does not stably exceed the strong baseline on all metrics; whereas when the two modules are used jointly, the performance is significantly improved. This indicates that DSEM and MARM modules are not two independent general enhancement modules that can separately bring stable gains, but complementary components designed for different bottlenecks.

First, compared with the Swin baseline, introducing DSEM alone improves $\mathrm{R}@5$ and $\mathrm{R}_{512}@5$ from 90.9\% and 97.7\% to 91.7\% and 98.2\%, respectively, while reducing the average localization error from 26.5~m to 25.5~m. This result indicates that enhancing directional structures and low-frequency spatial information improves recall in the candidate retrieval stage. Directional trends, regional contour relationships, and coarse-grained spatial layouts can provide more stable cross-modal matching cues than local textures. However, the $\mathrm{R}@1$ of DSEM decreases from 84.1\% to 82.8\%, indicating that structure enhancement alone is insufficient to support more precise top-1 retrieval. DSEM mainly addresses which information should be preserved and enhanced, but a distribution shift remains between the structure-enhanced thermal and satellite image features. Therefore, the enhanced structural information has not yet formed a sufficiently consistent cross-modal representation.

When MARM is introduced alone, the model achieves $\mathrm{R}@1=83.3\%$ and $\mathrm{R}@5=91.1\%$. Although $\mathrm{R}@5$ is slightly improved over the baseline, $\mathrm{R}@1$ remains lower and the average localization error increases to 29.4~m. This result shows that modality-aware statistical correction alone cannot provide stable performance gains. MARM is more suitable for remapping intermediate representations that already contain strong cross-modal stable cues. When directly applied to the original backbone features, the corrected representations still contain mixed semantic responses, local textures, noise, and modality biases, making the alignment target unclear. Although affine remapping can adjust the overall feature distribution, it cannot ensure that the preserved information corresponds to the cross-modal stable cues that are useful for retrieval.

When DSEM and MARM are used jointly, the model achieves substantial improvements. $\mathrm{R}@1$, $\mathrm{R}@5$, $\mathrm{R}_{512}@1$, and $\mathrm{R}_{512}@5$ reach 93.6\%, 97.4\%, 98.2\%, and 99.6\%, respectively, while the average localization error decreases to 12.3~m. These results show that DSEM and MARM form a clear sequential and complementary process. DSEM highlights relatively stable directional structures and low-frequency spatial layouts from the original features, shifting the representation from appearance-dominated to structure-dominated. MARM then performs modality-aware statistical correction on the structure-enhanced features, improving the consistency of thermal and satellite representations with respect to the same structural cues.


DSEM consists of three branches: the semantic-preserving branch, the directional response branch, and the low-frequency structure branch. To further analyze the role of its key internal components, we remove the directional response branch, the semantic-preserving branch, and the low-frequency structure branch under the full DSEM+MARM configuration, respectively. The experimental results are shown in Table~\ref{tab:ablation_branches}. As can be seen from Table~\ref{tab:ablation_branches}, all three branches contribute to the final performance. After removing the directional response branch, \(\mathrm{R}\text{@}1\) decreases from \(93.6\%\) to \(87.8\%\), indicating that directional structure modeling helps improve the discriminative capability of cross-modal retrieval. After removing the semantic-preserving branch, \(\mathrm{R}\text{@}1\) further decreases to \(81.5\%\), showing that preserving the contextual and discriminative information in the backbone features is also important. In comparison, removing the low-frequency structure branch causes the most significant performance degradation, with \(\mathrm{R}\text{@}1\) decreasing to \(70.5\%\) and \(\mathrm{R}_{512}\text{@}1\) decreasing to \(85.9\%\). This indicates that low-frequency spatial organization plays an important role in cross-modal matching under weak-texture and low-semantic scenes.

\begin{table}[!t]
\caption{Ablation results of the three internal branches of DSEM.
In the table, sem, dir, and low denote the semantic-preserving branch, directional response branch, and low-frequency structure branch of DSEM, respectively.
The best results are highlighted in \textbf{bold}.}
\label{tab:ablation_branches}
\centering
\footnotesize
\renewcommand{\arraystretch}{1.12}
\begin{tabular*}{0.98\columnwidth}{@{\hspace{4pt}}l@{\extracolsep{\fill}}cccccc@{\hspace{4pt}}}
\toprule
Backbone
& \multicolumn{3}{c}{DSEM}
& \(\mathrm{R}\text{@}1\)
& \(\mathrm{R}_{512}\text{@}1\)
& \(L_{512}^{2}\,(\mathrm{m})\) \\
\cmidrule(l{0pt}r{0pt}){2-4}
& sem & dir & low &  &  &  \\
\midrule
Swin-T & \(\checkmark\) & \(\checkmark\) & \(\checkmark\) & \textbf{93.6} & \textbf{98.2} & \textbf{12.3} \\
Swin-T & \(\checkmark\) &              & \(\checkmark\) & 87.8 & 95.2 & 14.3 \\
Swin-T &              & \(\checkmark\) & \(\checkmark\) & 81.5 & 91.8 & 16.9 \\
Swin-T & \(\checkmark\) & \(\checkmark\) &              & 70.5 & 85.9 & 22.4 \\
\bottomrule
\end{tabular*}
\end{table}

\subsection{Parameter Sensitivity Analysis}
To evaluate the sensitivity of the proposed method to key parameter choices, we analyze the convolution kernel size of the low-frequency structure branch in DSEM and the number of directions in the directional response branch.

\emph{Kernel size of the low-frequency structure branch.}
The low-frequency structure branch captures large-scale slowly varying terrain patterns through average pooling and large-kernel depthwise convolution, and the convolution kernel size directly determines its effective receptive field. As shown in Table~\ref{tab:low_kernel}, when the kernel size is \(3\), the receptive field is relatively small and mainly captures local details, resulting in insufficient modeling of large-scale terrain trends, with \(\mathrm{R}\text{@}1\) reaching only \(82.2\%\). When the kernel size is increased to \(7\), the receptive field is enlarged, but excessive smoothing leads to the loss of local discriminative information, causing \(\mathrm{R}\text{@}1\) to drop to \(85.5\%\). When the kernel size is \(5\), the best balance between the two is achieved, with \(\mathrm{R}\text{@}1\) reaching \(93.6\%\). Therefore, it is selected as the main configuration.

\begin{table}[!t]
\caption{Parameter sensitivity analysis of the convolution kernel size in the low-frequency structure branch of DSEM.
The best results are highlighted in \textbf{bold}.}
\label{tab:low_kernel}
\centering
\footnotesize
\renewcommand{\arraystretch}{1.12}
\begin{tabular*}{0.82\columnwidth}{@{\hspace{4pt}\extracolsep{\fill}}lcccc@{\hspace{4pt}}}
\toprule
Backbone 
& Kernel 
& \(\mathrm{R}\text{@}1\) 
& \(\mathrm{R}_{512}\text{@}1\) 
& \(L_{512}^{2}\,(\mathrm{m})\) \\
\midrule
Swin-T & 3 & 82.2 & 92.5 & 19.2 \\
Swin-T & 5 & \textbf{93.6} & \textbf{98.2} & \textbf{12.3} \\
Swin-T & 7 & 85.5 & 94.2 & 13.0 \\
\bottomrule
\end{tabular*}
\end{table}

\emph{Number of direction-sensitive units.}
The number of direction-sensitive convolutional units inside the directional response branch determines the coverage granularity of the directional space. As shown in Table~\ref{tab:direction_number}, when \(4\) direction-sensitive convolutional units are used, the compass direction coverage is relatively sparse, with \(\mathrm{R}\text{@}1\) of \(87.4\%\) and \(\mathrm{R}_{512}\text{@}1\) of \(94.8\%\). When expanded to \(8\) directions, the model can more uniformly cover the eight directions of N, NE, E, SE, S, SW, W, and NW, and the dominant structural direction at each spatial location can be perceived and distinguished in a more fine-grained manner, improving \(\mathrm{R}\text{@}1\) to \(93.6\%\). When further expanded to \(12\) directions, \(\mathrm{R}\text{@}1\) and \(\mathrm{R}_{512}\text{@}1\) are \(90.5\%\) and \(96.7\%\), respectively, which are lower than those of the \(8\)-direction configuration. However, it is worth noting that the \(L_{512}^{2}\) metric reaches the lowest value of \(11.5~\mathrm{m}\) under the \(12\)-direction configuration, outperforming the \(12.3~\mathrm{m}\) of the \(8\)-direction configuration. This is because the \(12\)-direction configuration covers a denser directional space, and for the retrieved Top-1 candidate, fine-grained modeling of structural directions helps further precise localization among neighboring candidates, thereby reducing the average localization error. However, increasing the number of directions also introduces higher parameter complexity and feature redundancy, causing the discriminative capability of the model for precise Top-1 matching \(\mathrm{R}\text{@}1\) to decrease instead. Excessive subdivision of the directional space leads to insufficient semantic distinction among some directional responses, weakening the ranking capability of global retrieval. Therefore, the \(8\)-direction configuration is selected as the main configuration. The advantage of the \(12\)-direction configuration on the \(L_{512}^{2}\) metric suggests that further increasing directional resolution may be worth exploring in fine-grained localization scenarios with coarse prior constraints.

\begin{table}[!t]
\caption{Parameter sensitivity analysis of the number of direction-sensitive units in DSEM.
The best results are highlighted in \textbf{bold}.}
\label{tab:direction_number}
\centering
\footnotesize
\renewcommand{\arraystretch}{1.12}
\begin{tabular*}{0.82\columnwidth}{@{\hspace{4pt}\extracolsep{\fill}}lcccc@{\hspace{4pt}}}
\toprule
Backbone 
& Num. Dir. 
& \(\mathrm{R}\text{@}1\) 
& \(\mathrm{R}_{512}\text{@}1\) 
& \(L_{512}^{2}\,(\mathrm{m})\) \\
\midrule
Swin-T & 4  & 87.4 & 94.8 & 13.4 \\
Swin-T & 8  & \textbf{93.6} & \textbf{98.2} & 12.3 \\
Swin-T & 12 & 90.5 & 96.7 & \textbf{11.5} \\
\bottomrule
\end{tabular*}
\end{table}

\subsection{Visualization Analysis}
\subsubsection{Stage-wise Grad-CAM Feature Response Analysis}
\begin{figure}[!t] 
\centering
\includegraphics[width=\linewidth]{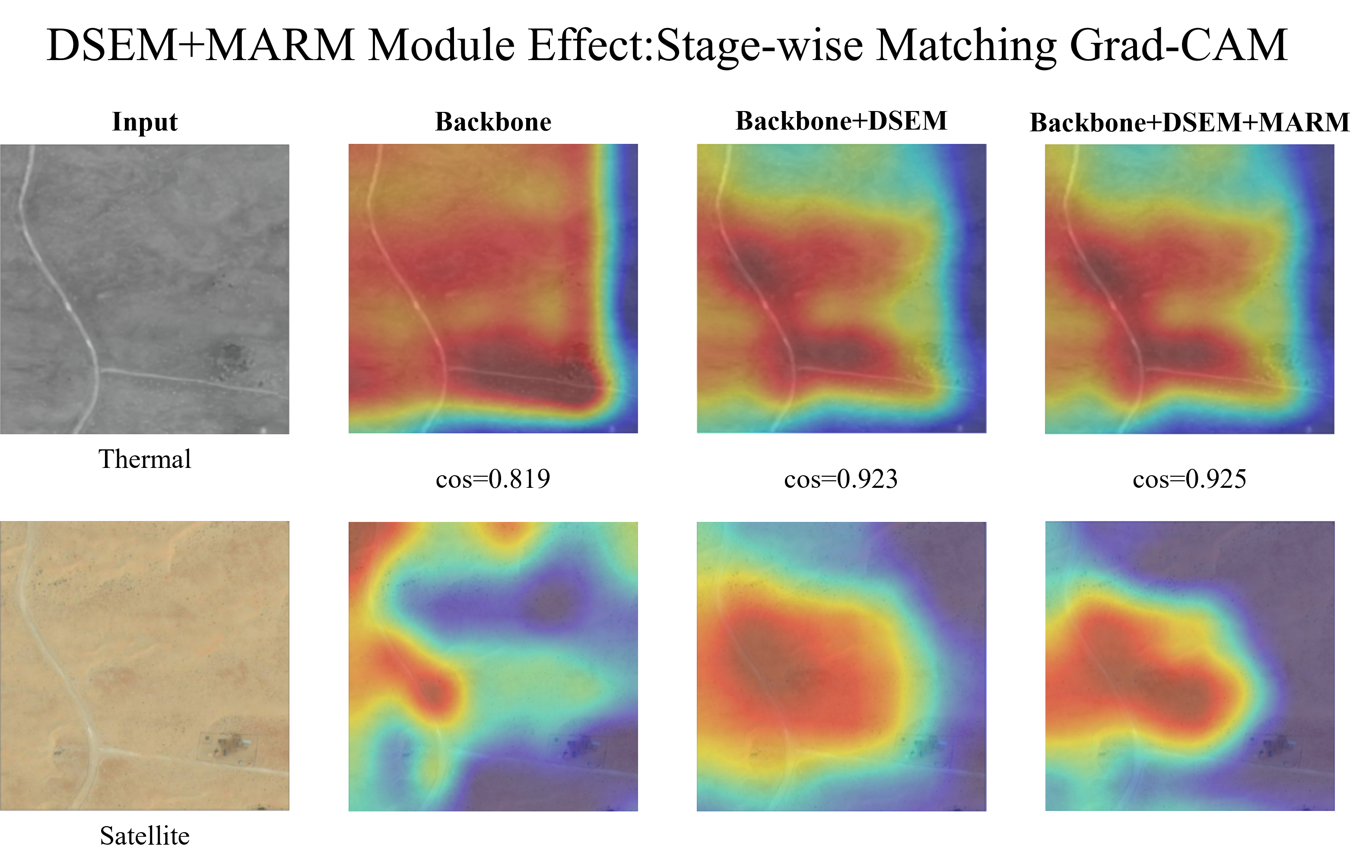}
\caption{Stage-wise Grad-CAM visualization analysis of the DSEM and MARM modules. The first column shows the original input images, with the thermal query image in the upper row and the corresponding satellite image in the lower row. The remaining three columns show the Grad-CAM heatmaps generated from the Swin-T backbone feature \(F\), the DSEM-enhanced feature \(F_{\mathrm{DSEM}}\), and the MARM remapped feature \(F_{\mathrm{MARM}}\), respectively. The ``cos'' value between the two images in each column denotes the cosine similarity between the global descriptors of the thermal query image and the satellite image at the corresponding stage. A higher value indicates stronger matching consistency between the two images in the current feature space. Warm-colored regions represent spatial locations that contribute more to the current matching similarity.}
\label{fig:stage_gradcam} 
\end{figure}

To intuitively evaluate the stage-wise influence of DSEM and MARM modules on cross-modal feature representation, we adopt Grad-CAM to generate heatmaps for the backbone output feature \(F\), the DSEM structure-enhanced feature \(F_{\mathrm{DSEM}}\), and the MARM remapped feature \(F_{\mathrm{MARM}}\), respectively. Meanwhile, we calculate the cosine similarity between the global descriptors of the thermal query image and the corresponding satellite image at each stage to quantify the cross-modal matching consistency of the feature representations at different stages. The visualization results are shown in Fig.~\ref{fig:stage_gradcam}.

As shown in Fig.~\ref{fig:stage_gradcam}, when only the backbone is used, the high-response regions of the query image and the satellite image are relatively scattered, and the spatial correspondence between the two images is inconsistent. The cosine similarity between the two images is 0.819, indicating that cross-modal matching based only on generic backbone features relies on relatively coarse evidence. After introducing DSEM, the high-response regions of the query image shift from broad background activation to more concentrated structural regions, while the responses of the satellite image are also reorganized from scattered local activations to regions related to the ground-object layout and directional trends. The spatial structural consistency between the two rows of heatmaps is clearly improved, and the cosine similarity increases to 0.923. This indicates that DSEM effectively highlights more stable shared matching cues under cross-modal conditions by strengthening directional structures and low-frequency spatial organizations. After further passing through MARM, the main response regions of the query image remain stable, while the response distribution of the satellite image becomes further concentrated toward the central region corresponding to the query structure, and the edge background noise is weakened. The cosine similarity between the two images reaches 0.925, indicating that MARM performs lightweight calibration on the feature distributions of the two modalities without destroying the existing structural matching evidence produced by DSEM, making the cross-modal matching evidence more coordinated and consistent.

\subsubsection{Retrieval Result Visualization Comparison}
\begin{figure}[!t] 
\centering
\includegraphics[width=\linewidth]{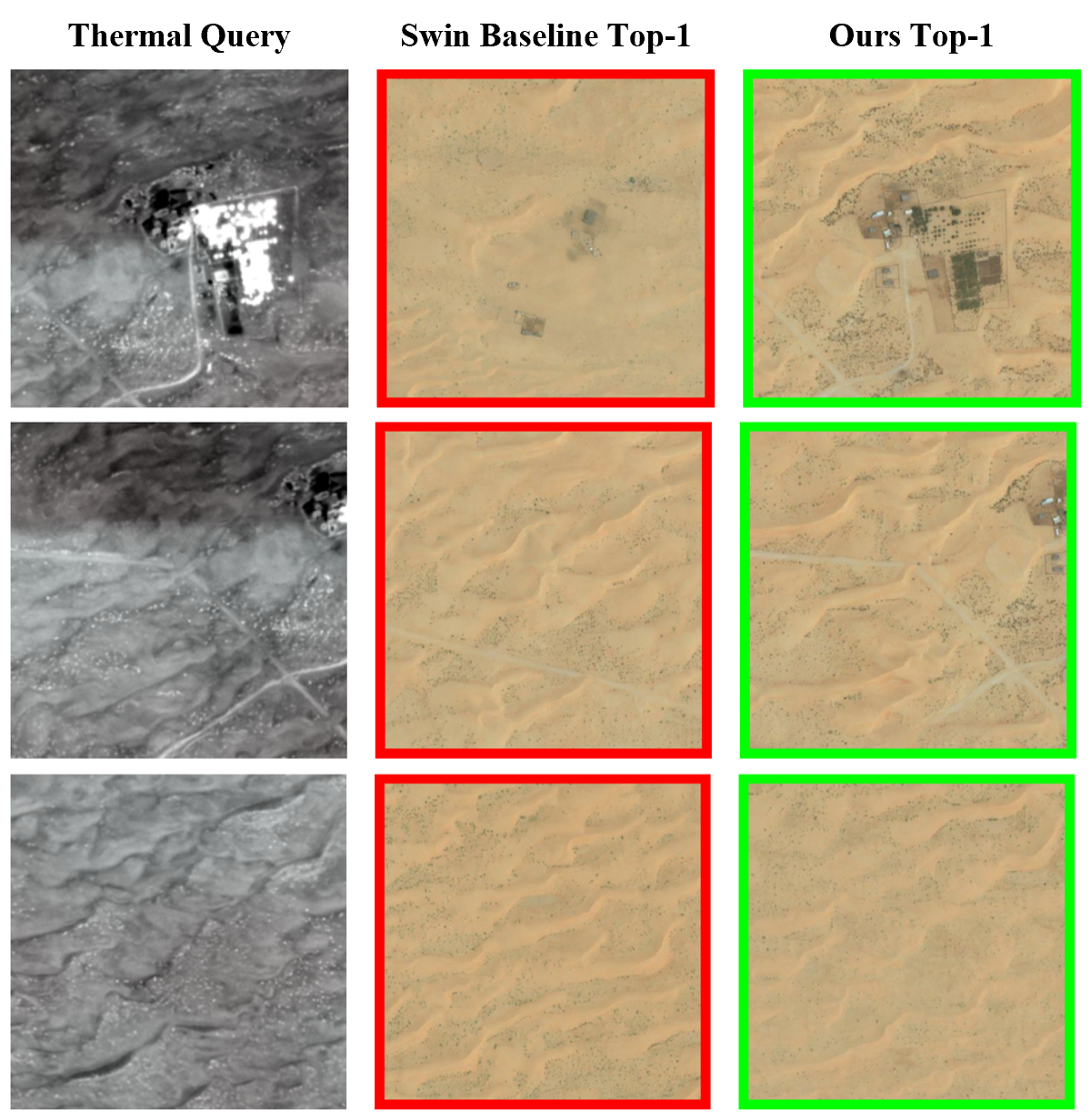}
\caption{Visualization comparison of retrieval results on the test set. Each row represents a test sample. The first column displays the UAV thermal query image; the second and third columns show the Top-1 satellite images retrieved by the Swin baseline (red box) and the proposed method (green box), respectively. A red box indicates a mismatch with the ground-truth location, while a green box denotes a correct match. As observed, in low-semantic scenes such as desert stripes and indistinct roads, the baseline method is prone to mismatches due to appearance inconsistencies, often retrieving images with similar landforms but incorrect locations. In contrast, the proposed method achieves more stable cross-modal matching by leveraging directional structures and low-frequency spatial organization.}
\label{fig:retrieval_results} 
\end{figure}

To further intuitively analyze the role of the proposed method, we visualize the retrieval results of several test samples, as shown in Fig.~\ref{fig:retrieval_results}. Compared with the baseline method, our method can more stably retrieve the satellite images corresponding to the geographic locations of the queries in weak-texture and low-semantic scenes, and ranks the correct matches at higher positions.

As can be seen from the results in Fig.~\ref{fig:retrieval_results}, when the scene lacks prominent building targets or regular road structures, the baseline method is more easily affected by appearance differences and inconsistencies in local textures, resulting in mismatched samples whose landforms are similar but whose geographic locations do not correspond. In contrast, our method exhibits better matching stability on samples with clear desert stripes, weak road extension directions, and coarse-grained regional layouts. This indicates that, between thermal UAV images and  satellite images, directional structures and low-frequency spatial organization are more suitable than color and local texture as the basis for cross-modal matching.

Combined with the ablation results, it can be further inferred that DSEM mainly functions by strengthening such more stable cross-modal structural cues, while MARM helps reduce the statistical distribution shift between thermal and satellite image features. Therefore, the complete method performs more stably in retrieval ranking and can more effectively distinguish true geographically corresponding samples from distractor samples that are only similar in local appearance. 


\section{Conclusions}
\label{sec:conclusions}

This paper investigates UAV geo-localization in GNSS-denied and low-visibility conditions through cross-modal matching between  UAV thermal and satellite imagery. The presence of cross-modal imaging discrepancies and  cross-view variations severely compromises the ability of appearance-based methods to establish stable correspondences. To tackle this, we propose SAMF, a novel framework that explicitly exploit and utilizes stable structural cues shared between thermal and satellite images to improve the accuracy and reliability of   UAV thermal geo-localization. SAMF incorporates a directional structure enhancement module to extract stable cross-modal structural cues by reinforcing local directional responses and low-frequency spatial organization. Concurrently, a modality-aware affine remapping module applies modality-aware statistical correction to these enhanced features, further mitigating cross-modal distribution shifts. We have also introduced the Thermal-UAV-Paired evaluation benchmark derived from the Thermal-UAV dataset. Comprehensive evaluations on the Boson-nighttime and Thermal-UAV-Paired datasets reveal that SAMF achieves superior performance over the original Swin baseline, state-of-the-art  UAV thermal geo-localization methods, and representative thermal-visible cross-modal retrieval approaches.


\section*{Acknowledgments}
This work is partially supported by National Key Laboratory of Space Datum Fund Project under grant No. SKLSD2025-ZZ-14, National Key Laboratory of Integrated Aircraft Control Technology Fund Project under grant No. WDZC2024601B02, Major Science and Technology Projects of Henan Province under grant No.251100240100, and Key Project of Geographic Science and Engineering Department of Henan University under grant No.DLXKKY250308.

 
\vspace{11pt}

\begin{IEEEbiography}[{\includegraphics[width=1in,height=1.25in,clip,keepaspectratio]{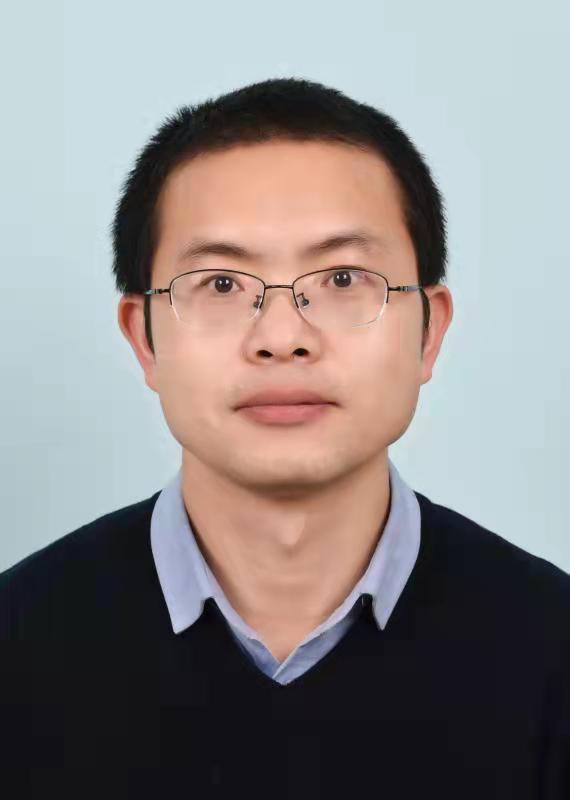}}]{Chun Liu}
 received the Ph.D. from Chinese Academy of Sciences in 2012. Currently, he is a professor at Henan University, and serves on the Intelligent Surveying and Mapping Working Committee of the China Surveying and Mapping Society. His research focuses on spatial intelligent and dependability computing.
\end{IEEEbiography}


\vspace{-1.5\baselineskip}

\begin{IEEEbiography}[{\includegraphics[width=1in,height=1.25in,clip,keepaspectratio]{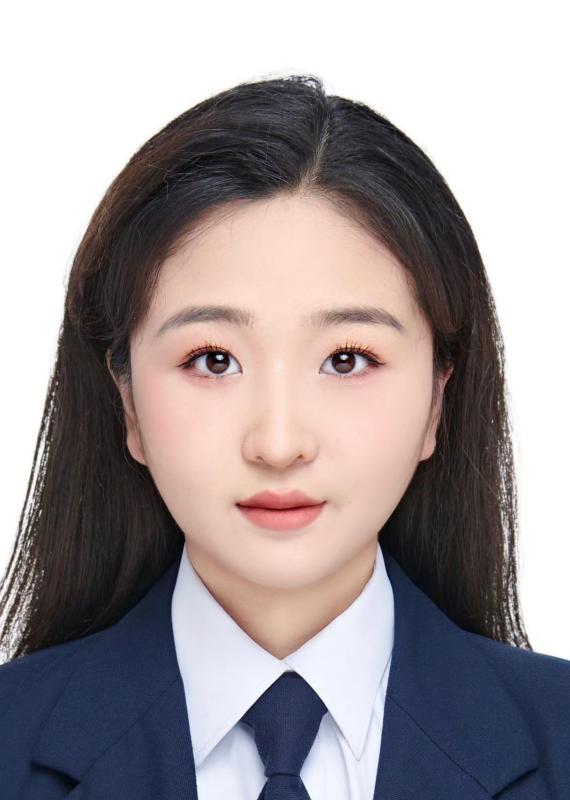}}]{Yuxin Wang}
received the B.S. degree from Jiangxi Agricultural University, China, in 2024. She is currently pursuing the M.S. degree with Henan University, China. Her research interests include computer vision, UAV geo-localization, and cross-modal image matching. 
\end{IEEEbiography}
\vspace{-1.5\baselineskip}

\begin{IEEEbiography}[{\includegraphics[width=1in,height=1.25in,clip,keepaspectratio]{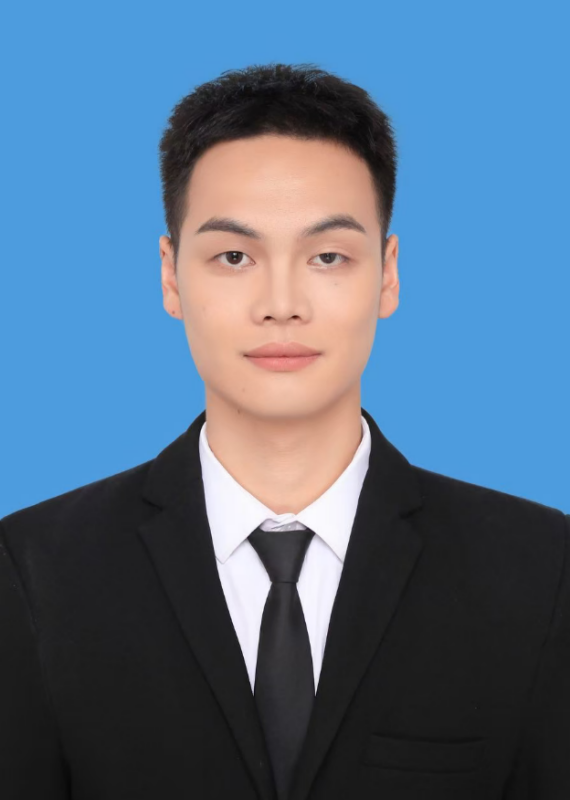}}]{Zhenxiang Huo}
received the M.S. degree  from Henan University in 2025. He is currently pursuing the Ph.D. degree at Henan University. His main research interests include computer vision and UAV cross-view geo-localization. 
\end{IEEEbiography}
\vspace{-1.5\baselineskip}

\begin{IEEEbiography}[{\includegraphics[width=1in,height=1.25in,clip,keepaspectratio]{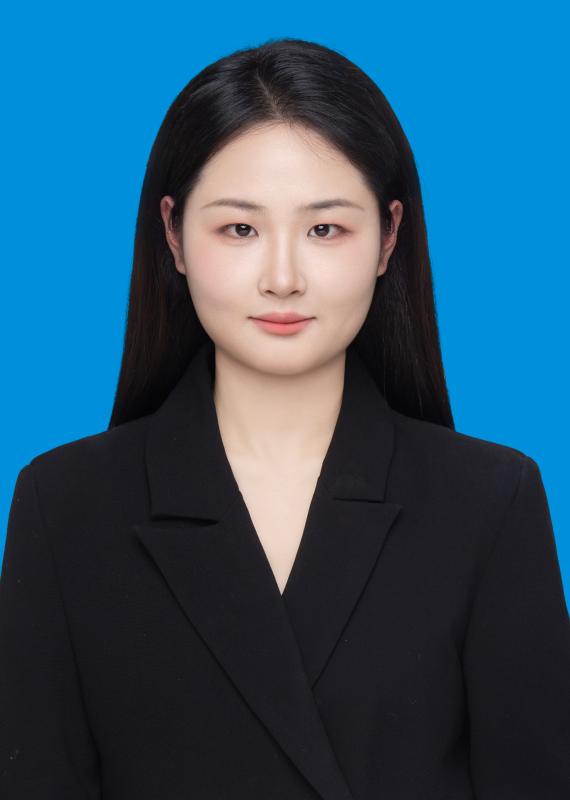}}]{Linlin Zhu} received the Ph.D. degree in Computer Science and Technology from Xi’an Jiaotong University in 2026. She is currently an assistant Professor at Henan University. Her research interests include artificial intelligence and multimodal sentiment analysis, and multimodal visual geo-localization.

\end{IEEEbiography}
\vspace{-1.5\baselineskip}

\begin{IEEEbiography}[{\includegraphics[width=1in,height=1.25in,clip,keepaspectratio]{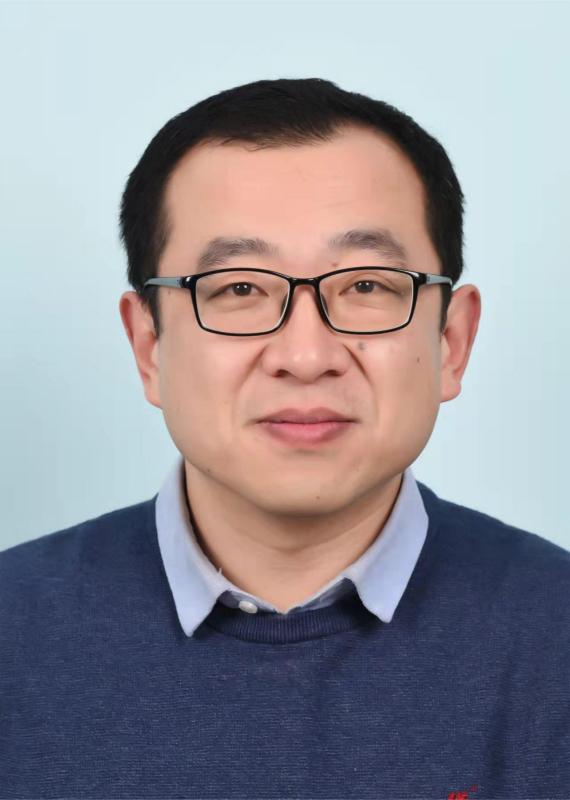}}]{Tao Xu}
received the Ph.D. degree from East China Normal University in 2017. He is currently an Associate Professor at Henan University, Zhengzhou, China. His research interests include image geo-localization, graph-based spatiotemporal prediction, geospatial artificial intelligence, and inference and applications of multimodal large models.
\end{IEEEbiography}
\vspace{-1.5\baselineskip}

\begin{IEEEbiography}[{\includegraphics[width=1in,height=1.25in,clip,keepaspectratio]{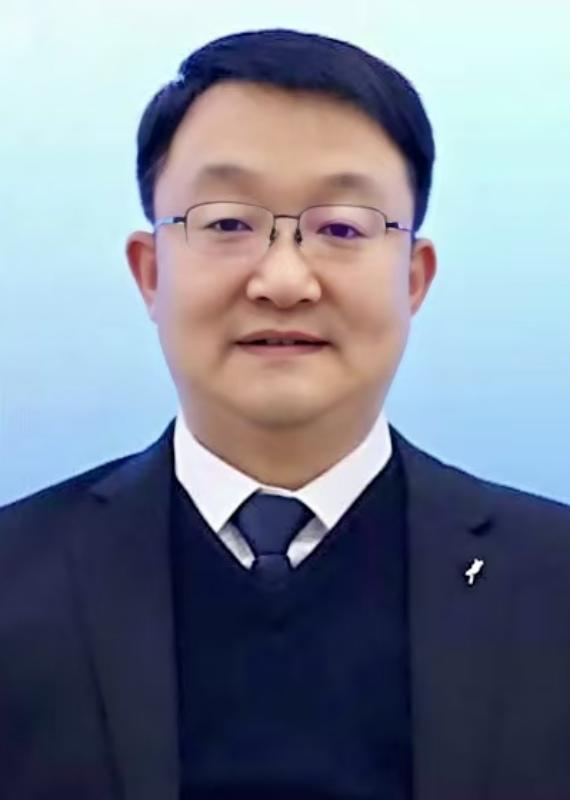}}]{Henggang Han}
is the deputy director of the low-altitude airspace management R\&D center of Lantian Laboratory, and deputy general manager of Henan Yunhuan UAV Network Technology Co., Ltd., China.  His work focuses on UAVs and the low-altitude economy.
\end{IEEEbiography}

\vspace{11pt}


\vfill


\begin{thebibliography}{47}
\bibliographystyle{IEEEtran}

\bibitem{ref1}
A. Couturier and M. A. Akhloufi, “A review on absolute visual localization for UAV,” \textit{Robot. Auton. Syst.}, vol. 135, p. 103666, 2021.

\bibitem{ref2}
S. Weiss, M. W. Achtelik, S. Lynen, M. C. Achtelik, L. Kneip, M. Chli, et al., “Monocular Vision for Long-term Micro Aerial Vehicle State Estimation: A Compendium,” \textit{J. Field Robot.}, vol. 30, no. 5, pp. 803–831, 2013.

\bibitem{ref3}
H. Goforth and S. Lucey, “GPS-Denied UAV Localization using Pre-existing Satellite Imagery,” in \textit{Proc. IEEE Int. Conf. Robot. Autom. (ICRA)}, 2019, pp. 2974–2980.

\bibitem{ref4}
J. Xiao, D. Tortei, E. Roura, and G. Loianno, “Long-Range UAV Thermal Geo-Localization with Satellite Imagery,” in \textit{Proc. IEEE/RSJ Int. Conf. Intell. Robots Syst. (IROS)}, 2023, pp. 5820–5827.

\bibitem{ref5}
S. Workman, R. Souvenir, and N. Jacobs, “Wide-Area Image Geolocalization with Aerial Reference Imagery,” in \textit{Proc. IEEE Int. Conf. Comput. Vis. (ICCV)}, 2015, pp. 3961–3969.

\bibitem{ref6}
Z. Zheng, Y. Wei, and Y. Yang, “University-1652: A Multi-view Multisource Benchmark for Drone-based Geo-localization,” in \textit{Proc. 28th ACM Int. Conf. Multimedia (ACM MM)}, 2020, pp. 1395–1403.

\bibitem{ref7}
G. Berton, et al., “Deep Visual Geo-localization Benchmark,” in \textit{Proc. IEEE/CVF Conf. Comput. Vis. Pattern Recognit. (CVPR)}, 2022, pp. 5386–5397.

\bibitem{ref8}
R. Arandjelovic, P. Gronat, A. Torii, T. Pajdla, and J. Sivic, “NetVLAD: CNN Architecture for Weakly Supervised Place Recognition,” in \textit{Proc. IEEE Conf. Comput. Vis. Pattern Recognit. (CVPR)}, 2016, pp. 5297–5307.

\bibitem{ref9}
S. Hausler, S. Garg, M. Xu, M. Milford, and T. Fischer, “PatchNetVLAD: Multi-Scale Fusion of Locally-Global Descriptors for Place Recognition,” in \textit{Proc. IEEE/CVF Conf. Comput. Vis. Pattern Recognit. (CVPR)}, 2021, pp. 14136–14147.

\bibitem{ref10}
Y. Ganin and V. Lempitsky, “Unsupervised Domain Adaptation by Backpropagation,” in \textit{Proc. Int. Conf. Mach. Learn. (ICML)}, 2015, pp. 1180–1189.

\bibitem{ref11}
P. Isola, J.-Y. Zhu, T. Zhou, and A. A. Efros, “Image-to-Image Translation with Conditional Adversarial Networks,” in \textit{Proc. IEEE Conf. Comput. Vis. Pattern Recognit. (CVPR)}, 2017, pp. 5967–5976.

\bibitem{ref12}
X. Qian, M. Zhang, and F. Zhang, “Sparse GANs for Thermal Infrared Image Generation From Optical Image,” \textit{IEEE Access}, vol. 8, pp. 180124–180132, 2020.

\bibitem{ref13}
J. Xiao, N. Zhang, D. Tortei, and G. Loianno, “STHN: Deep Homography Estimation for UAV Thermal Geo-Localization With Satellite Imagery,” \textit{IEEE Robot. Autom. Lett.}, vol. 9, no. 10, pp. 8754–8761, 2024.

\bibitem{ref14}
J. Xiao and G. Loianno, “UASTHN: Uncertainty-Aware Deep Homography Estimation for UAV Satellite-Thermal Geo-Localization,” in \textit{Proc. IEEE Int. Conf. Robot. Autom. (ICRA)}, 2025, pp. 14066–14072.

\bibitem{ref15}
X. Li, G. Zhang, H. Cui, S. Hou, Y. Chen, Z. Li, et al., “Progressive fusion learning: A multimodal joint segmentation framework for building extraction from optical and SAR images,” \textit{ISPRS J. Photogramm. Remote Sens.}, vol. 195, pp. 178–191, 2023.

\bibitem{ref16}
J. Chen, J. Zhu, P. He, Y. Guo, L. Hong, Y. Yang, et al., “Unsupervised Domain Adaptation for Building Extraction of High-Resolution Remote Sensing Imagery Based on Decoupling Style and Semantic Features,” \textit{IEEE Trans. Geosci. Remote Sens.}, vol. 62, pp. 1–17, 2024.

\bibitem{ref17}
X. Zhang, Y. Liu, J. Liang, K. Li, Z. Huang, and H. Xiao, “SCC-Loc: A Unified Semantic Cascade Consensus Framework for UAV Thermal Geo-Localization,” \textit{IEEE J. Sel. Top. Appl. Earth Obs. Remote Sens.}, vol. 19, pp. 1–17, 2026.

\bibitem{ref18}
A. Shetty and G. X. Gao, “UAV Pose Estimation using Cross-view Geolocalization with Satellite Imagery,” in \textit{Proc. IEEE Int. Conf. Robot. Autom. (ICRA)}, 2019, pp. 1827–1833.

\bibitem{ref19}
X. Tian, J. Shao, D. Ouyang, and H. T. Shen, “UAV-Satellite View Synthesis for Cross-View Geo-Localization,” \textit{IEEE Trans. Circuits Syst. Video Technol.}, vol. 32, no. 7, pp. 4804–4815, 2022.

\bibitem{ref20}
T. Wang, Z. Zheng, C. Yan, J. Zhang, Y. Sun, B. Zheng, et al., “Each Part Matters: Local Patterns Facilitate Cross-View Geo-Localization,” \textit{IEEE Trans. Circuits Syst. Video Technol.}, vol. 32, no. 2, pp. 867–879, 2022.

\bibitem{ref21}
M. Dai, J. Hu, J. Zhuang, and E. Zheng, “A Transformer-Based Feature Segmentation and Region Alignment Method for UAV-View Geo-Localization,” \textit{IEEE Trans. Circuits Syst. Video Technol.}, vol. 32, no. 7, pp. 4376–4389, 2022.

\bibitem{ref22}
Z. Cui, P. Zhou, X. Wang, Z. Zhang, Y. Li, H. Li, et al., “A Novel Geo-Localization Method for UAV and Satellite Images Using Cross-View Consistent Attention,” \textit{Remote Sens.}, vol. 15, no. 19, p. 4667, 2023.

\bibitem{ref23}
M. Bianchi and T. D. Barfoot, “UAV Localization Using Autoencoded Satellite Images,” \textit{IEEE Robot. Autom. Lett.}, vol. 6, no. 2, pp. 1761–1768, 2021.

\bibitem{ref24}
F. Deuser, K. Habel, and N. Oswald, “Sample4Geo: Hard Negative Sampling For Cross-View Geo-Localisation,” in \textit{Proc. IEEE/CVF Int. Conf. Comput. Vis. (ICCV)}, 2023, pp. 16801–16810.

\bibitem{ref25}
R. Zhu, L. Yin, M. Yang, F. Wu, Y. Yang, and W. Hu, “SUES-200: A Multi-Height Multi-Scene Cross-View Image Benchmark Across Drone and Satellite,” \textit{IEEE Trans. Circuits Syst. Video Technol.}, vol. 33, no. 9, pp. 4825–4839, 2023.

\bibitem{ref26}
J. Fan, E. Zheng, Y. He, and J. Yang, “A Cross-View Geo-Localization Algorithm Using UAV Image and Satellite Image,” \textit{Sensors}, vol. 24, no. 12, p. 3719, Jun. 7 2024.

\bibitem{ref27}
J. Hoffman, et al., “CyCADA: Cycle-Consistent Adversarial Domain Adaptation,” in \textit{Proc. Int. Conf. Mach. Learn. (ICML)}, 2018, pp. 1989–1998. [Online].

\bibitem{ref28}
J.-Y. Zhu, T. Park, P. Isola, and A. A. Efros, “Unpaired Image-to-Image Translation Using Cycle-Consistent Adversarial Networks,” in \textit{Proc. IEEE Int. Conf. Comput. Vis. (ICCV)}, 2017, pp. 2242–2251.

\bibitem{ref29}
X. Huang and S. Belongie, “Arbitrary Style Transfer in Real-Time with Adaptive Instance Normalization,” in \textit{Proc. IEEE Int. Conf. Comput. Vis. (ICCV)}, 2017, pp. 1510–1519.

\bibitem{ref30}
H. Nam and H.-E. Kim, “Batch-Instance Normalization for Adaptively Style-Invariant Neural Networks,” in \textit{Advances in Neural Information Processing Systems}, S. Bengio, H. Wallach, H. Larochelle, K. Grauman, N. Cesa-Bianchi, and R. Garnett, Eds. Curran Associates, Inc., 2018 [Online].

\bibitem{ref31}
F. Liu, C. Gao, Y. Sun, Y. Zhao, F. Yang, A. Qin, et al., “Infrared and Visible Cross-Modal Image Retrieval Through Shared Features,” \textit{IEEE Trans. Circuits Syst. Video Technol.}, vol. 31, no. 11, pp. 4485–4496, 2021.

\bibitem{ref32}
L. Liao, M. Yang, and B. Zhang, “Deep Supervised Dual Cycle Adversarial Network for Cross-Modal Retrieval,” \textit{IEEE Trans. Circuits Syst. Video Technol.}, vol. 33, no. 2, pp. 920–934, 2023.

\bibitem{ref33}
Q. Cheng, Z. Tan, K. Wen, C. Chen, and X. Gu, “Semantic Pre-Alignment and Ranking Learning With Unified Framework for Cross-Modal Retrieval,” \textit{IEEE Trans. Circuits Syst. Video Technol.}, vol. 34, no. 7, pp. 6503–6516, 2024.

\bibitem{ref34}
T. Liang, Y. Jin, W. Liu, T. Wang, S. Feng, and Y. Li, “Bridging the Gap: Multi-Level Cross-Modality Joint Alignment for Visible-Infrared Person Re-Identification,” IEEE Trans. Circ. Syst. Video Tech., vol. 34, no. 8, pp. 7683–7698, 2024.

\bibitem{ref35}
K. Ren and L. Zhang, “Implicit Discriminative Knowledge Learning for Visible-Infrared Person Re-Identification,” in Proceedings of the IEEE/CVF Conference on Computer Vision and Pattern Recognition (CVPR), Jun. 2024, pp. 393–402.

\bibitem{ref36}
X. Hua, K. Cheng, H. Lu, J. Tu, Y. Wang, and S. Wang, “MSCMNet: Multi-scale Semantic Correlation Mining for Visible-Infrared Person Re-Identification,” Pattern Recognit., vol. 159, p. 111090, 2025.

\bibitem{ref37}
H. Zhao, S. Yang, X. Li, J. Peng, and X. Wu, “Middle modality interactive feature attention learning for visible-infrared person re-identification,” Neurocomputing, vol. 660, no. C, p. 131815, Jun. 2026.

\bibitem{ref38}
P. Maheshwari, et al., “AnyThermal: Towards Learning Universal Representations for Thermal Perception,” in 2026 IEEE International Conference on Robotics and Automation (ICRA), 2026.

\bibitem{ref39}
R. A. Kirsch, “Computer determination of the constituent structure of biological images,” \textit{Comput. Biomed. Res.}, vol. 4, no. 3, pp. 315–328, Jun. 1971.

\bibitem{ref40}
Y. Liu, M. M. Cheng, X. Hu, J. W. Bian, L. Zhang, X. Bai, et al., “Richer Convolutional Features for Edge Detection,” \textit{IEEE Trans. Pattern Anal. Mach. Intell.}, vol. 41, no. 8, pp. 1939–1946, Aug. 2019.

\bibitem{ref41}
K. Zhao, Q. Han, C.-B. Zhang, J. Xu, and M.-M. Cheng, “Deep Hough Transform for Semantic Line Detection,” \textit{IEEE Trans. Pattern Anal. Mach. Intell.}, vol. 44, no. 9, pp. 4793–4806, Sep. 2022.

\bibitem{ref42}
J. Liu, X. Fan, J. Jiang, R. Liu, and Z. Luo, “Learning a Deep Multi-Scale Feature Ensemble and an Edge-Attention Guidance for Image Fusion,” \textit{IEEE Trans. Circuits Syst. Video Technol.}, vol. 32, no. 1, pp. 105–119, 2022.

\bibitem{ref43}
K. Wang, Z. Wang, Z. Li, A. Su, X. Teng, E. Pan, et al., “Oriented object detection in optical remote sensing images using deep learning: A survey,” \textit{Artif. Intell. Rev.}, vol. 58, no. 11, p. 350, Aug. 2025.

\bibitem{ref44}
Z. Liu, et al., “Swin Transformer: Hierarchical Vision Transformer using Shifted Windows,” in \textit{Proc. IEEE/CVF Int. Conf. Comput. Vis. (ICCV)}, 2021, pp. 9992–10002.

\bibitem{ref45}
K. He, X. Zhang, S. Ren, and J. Sun, “Deep Residual Learning for Image Recognition,” in \textit{Proc. IEEE Conf. Comput. Vis. Pattern Recognit. (CVPR)}, 2016, pp. 770–778.

\bibitem{ref46}
W. Zhang, Z. Zhong, H. Chen, and N. Jing, “GeoViewMatch: A MultiScale Feature-Matching Network for Cross-View Geo-Localization Using Swin-Transformer and Contrastive Learning,” \textit{Remote Sens.}, vol. 16, no. 4, p. 678, 2024.

\bibitem{ref47}
J. Liu, K. Ren, and Q. Chen, “Swin transformer for feature extraction: A cross-view geo-localization method for UAV-Satellite views,” \textit{Int. J. Remote Sens.}, vol. 46, no. 13, pp. 4864–4882, 2025.

\end{thebibliography}
\end{document}